\documentclass[10pt, letterpaper, journal, compsoc]{IEEEtran}
\usepackage{algorithm}
\usepackage{multirow}
\usepackage{amsmath,amssymb,amsfonts}
\usepackage{amsthm}
\usepackage{graphicx}
\usepackage{setspace}
\usepackage{multicol}
\usepackage{multirow}
\usepackage{cite}
\usepackage{float}
\usepackage{color}
\usepackage{alltt, listings}
\usepackage{subfigure}
\usepackage{algpseudocode}
\usepackage{wrapfig}
\usepackage{booktabs} % for professional tables
\usepackage{url}
\usepackage{bbding}
\usepackage{bm}

\theoremstyle{definition}

\begin {document}

\title{Stabilizing and Improving Federated Learning with Non-IID Data and Client Dropout}
%\title{Bias-Variance Reduced }

%\author{
%	{\rm Jian Xu, Meiling Yang, Shao-Lun Huang} \\
%%	$^*$Tsinghua-Berkeley Shenzhen Institute, Tsinghua University\\
%%	xujian20@mails.tsinghua.edu.cn, shaolun.huang@sz.tsinghua.edu.cn
%%xx$^\ddagger$
%}
%\author{\IEEEauthorblockN{Anonymous Authors}}

\author{
	Jian~Xu,
	Meilin~Yang,
	Wenbo~Ding,~\IEEEmembership{Member,~IEEE,}
	and~Shao-Lun~Huang,~\IEEEmembership{Member,~IEEE}% <-this % stops a space
	\thanks{Manuscript received. The research of Shao-Lun Huang is supported in part by the Shenzhen Science and Technology Program under Grant KQTD20170810150821146, National Key R\&D Program of China under Grant 2021YFA0715202. \textit{(Corresponding author: Shao-Lun~Huang.)}}
	\IEEEcompsocitemizethanks{
		\IEEEcompsocthanksitem Jian~Xu, Meilin~Yang, Wenbo~Ding and Shao-Lun~Huang are with the Tsinghua-Berkeley Shenzhen Institute, Tsinghua Shenzhen International Graduate School, Tsinghua University, Shenzhen, China. E-mail:\{xujian20, yml21\}@mails.tsinghua.edu.cn; \{ding.wenbo, shaolun.huang\}@sz.tsinghua.edu.cn.
		% \protect\\
		% E-mail: see http://www.michaelshell.org/contact.html
%		\IEEEcompsocthanksitem 
%		\IEEEcompsocthanksitem
	}% <-this % stops a space
}

\IEEEtitleabstractindextext{
	
\begin{abstract}
	
	The label distribution skew induced data heterogeniety has been shown to be a significant obstacle that limits the model performance in federated learning, which is particularly developed for collaborative model training over decentralized data sources while preserving user privacy. This challenge could be more serious when the participating clients are in unstable circumstances and dropout frequently. Previous work and our empirical observations demonstrate that the classifier head for classification task is more sensitive to label skew and the unstable performance of FedAvg mainly lies in the imbalanced training samples across different classes. The biased classifier head will also impact the learning of feature representations. Therefore, maintaining a balanced classifier head is of significant importance for building a better global model. To this end, we propose a simple yet effective framework by introducing a prior-calibrated softmax function for computing the cross-entropy loss and a prototype-based feature augmentation scheme to re-balance the local training, which are lightweight for edge devices and can facilitate the global model aggregation. The improved model performance over existing baselines in the presence of non-IID data and client dropout is demonstrated by conducting extensive experiments on benchmark classification tasks.
	
\end{abstract}

}

\maketitle

\ifCLASSOPTIONcompsoc
\IEEEraisesectionheading{\section{Introduction}\label{sec:intro}}
\else
\section{Introduction}\label{sec:intro}
\fi

\IEEEPARstart{W}{ith} the rapid development of Internet of Things (IoT) and machine learning techniques, significant amounts of data are generated in every day and can benefit intelligent applications in a wide variety of scenarios, such as speech recognition, healthcare systems and smart cities \cite{hinton2012speech,Hay22Health,Ullah20smart}. However, collecting data from edge devices for centralized model training causes high communication cost and serious privacy concerns, i.e., some data may contain personal and sensitive information \cite{sharma2019data}. As a result, training deep learning models centrally by gathering data from user devices might be prohibited and impractical. To alleviate the above issues, it is attractive to adopt federated learning (FL) \cite{mcmahan17FL,yang2019federated,LiS20FLReview} to collectively train models, which can take full advantages of the local data and edge computational resources in the IoT systems. FL is a distributed computing paradigm where multiple edge devices, i.e., clients, collaboratively train a global model without disclosing local sensitive data. Conventional FL systems usually consist of a central server and a number of clients, where each client updates the model based on local data and the server is responsible for synchronizing model parameters. The primary goal of FL is to efficiently train a global model with the highest possible accuracy \cite{Li20FedProx,Waha21FMLSurvey}.

Despite the merits, FL in IoT systems could also meet with plenty of challenges due to potentially heterogeneous device capabilities and data distributions, which are also known as system heterogeneity and non-IID data \cite{Kairouz21AdvanceFL}. The non-IID data in FL means the underlying distributions of local data are not identical across clients \cite{zhao2018nonIID}. System heterogeneity could be caused by different resource constraints, including storage, computing capability, energy and network bandwidth \cite{Wang19adaFL}. In synchronous FL, aggregation is conducted when all the selected clients fully complete their local training and return their model updates. However, the clients could have significantly different performances in real-world scenarios, which implies that some faster clients have to wait for the slower ones, i.e., stragglers, resulting in unnecessary time-consuming. Moreover, some clients might be unresponsive due to inactive devices and/or communication failures and thus induce unavailability of some local updates, i.e., dropout. In this work, we focus on cross-device federated learning, where the participants are mostly edge devices such as smart phones and IoTs. These devices generally have limited computing and communication resources, and also have different training data distributions. Considering that federated learning typically takes a relatively large number of communication rounds to converge, it is difficult to ensure that all devices will be available or return back their local updates on time during the entire training process, leading to severe straggler and dropout issues \cite{Kairouz21AdvanceFL}. 

To alleviate the straggler issue, FL algorithms usually perform partial participation and intermittent model aggregation after multiple local epochs as proposed in \cite{mcmahan17FL}. In such settings, however, the non-IID data brings new challenges for model aggregation due to the high diversity of local models. As a result, the resulting global model could have poor and unstable performance during the training process \cite{zhao2018nonIID}. Moreover, it also takes more communication rounds for convergence and thus prolongs the on-clock training time. Therefore, developing FL algorithms with resilience to non-IID data and client dropout is of significant importance for fully taking advantages of scattered data generated from edge devices. Although tremendous efforts have been devoted to combating the non-IID data \cite{Karimireddy20SCAFFOLD,Acar21FedDyn,LiHS21MOON}, the performance is still far from being satisfactory. While many factors can lead to the data heterogeneity among clients, the label skew is demonstrated to be the main reason of model performance deterioration as it could lead to inconsistency between local objectives and global objective \cite{Li21FLNonIID}. On the other hand, previous works on centralized learning with long-tailed data and FL with class imbalance have demonstrated that the final decision layer is most vulnerable to class imbalance and will result in significant bias of decision boundaries \cite{Ren20BMS,Luo21CCVR}. To alleviate it, several methods including weight normalization of output layer and embedding augmentation of tail classes are investigated to rectify the model training for fair performance among all classes \cite{ZhuHZ21DataFreeDistill,Luo21CCVR}. Inspired by those observations, we aim to train balanced classifiers during the local training stage to facilitate the global aggregation.

In this paper, we focus on mitigating the label skew and client dropout issues simultaneously, with especial interest in the case where each client only contains samples from a few categories. We emphases that the partial participation of clients is not a decision by algorithms and is uncontrollable due to unexpected client dropout in our settings. We propose a novel \underline{FL} framework with \underline{Re}-\underline{Ba}lanced local classifier training ({ReBaFL}) that not only learns from the local data but also preserves the global knowledge. To be more specific, we propose a relaxed version of prior-calibrated softmax  function and a prototype-based feature augmentation scheme in local training to reduce the bias of local models, thus benefiting the aggregated global model. The proposed framework is evaluated on benchmark datasets with extensive experiments and the effectiveness in stabilizing the federated learning process is verified. Numerical results demonstrate that non-trivial improvements in test accuracy could be achieved, despite highly skewed label distributions across clients and severe client dropout. To sum up, this work makes the following contributions:
%\vspace{-1ex}
\begin{itemize}
	\item A novel FL framework with label prior calibrated local training is developed for alleviating catastrophic forgetting when clients only observe limited labels.
	
	\item A prototype-based feature augmentation scheme is employed to speed up the model convergence. Feature inverse attacks are also evaluated to indicate that using feature prototypes will not violate data privacy.
	
	\item The efficacy and superiority of our proposed method are verified on benchmark datasets with various levels of non-IID data and client dropout.
\end{itemize}

The rest of the paper is organized as follows. Section \ref{secRelated} reviews the related works. The background and motivation are introduced in Section \ref{sec:Prelim}. The design of our framework is elaborated and evaluated in Section \ref{sec:Framework} and Section \ref{secExperiment}, respectively. Finally, Section \ref{secConclusion} concludes this paper.

\section{Related Work}
\label{secRelated}

\subsection{Federated Learning with Non-IID Data}
The non-IID data issue has attained significant attentions since the emergence of FL. A number of studies focus on improving the optimization algorithms by designing various regularization terms \cite{Li20FedProx,Zhang2021FedPD,Acar21FedDyn,LiHS21MOON} and gradient correction schemes \cite{Karimireddy20SCAFFOLD,OzfaturaOG21FedADC,xu2021fedcm}. For instance, FedDP \cite{Zhang2021FedPD} leverages the augmented Lagrangian approach to seek consensus in local updates. FedCM \cite{xu2021fedcm} introduces a client-side momentum to debias the local gradient direction. Data re-sampling and re-weighting techniques could also be applied to seek a class-balanced model training thus improve the global performance \cite{HsuQ020FedIR,Wang21ImbalanceFL}. However, this type of approaches are invalid when samples from certain classes are totally missing. Besides, data sharing and augmentation methods are also extensively investigated to mitigate the non-IID data challenges \cite{zhao2018nonIID,YoonSHY21FedMix}. From the perspective of client selection, sampling clients with high contributions to model performance is able to speed up the convergence \cite{Wang20Favor,Tang21FedGP,FraboniVKL21ClusteredSampling,ChenC21FedBE}. However, the server either needs prior knowledge about the clients or should record the statistical behaviors of each client to decide the selection strategy, which could be impractical or induce additional computation costs to the server. When auxiliary data is available at the server or generative model is trained, knowledge distillation approaches can be employed to obtain a more refined global model \cite{LinKSJ20FedDF,ZhuHZ21DataFreeDistill,Zhang22FedFTG}. Another line of work attempts to overcome data heterogeneity by developing personalized FL \cite{kulkarni2020surveyPFL,Dinh20pFedMe,Fallah20pFedAvg,Collins21FedRep} or clustered FL \cite{ghosh2020efficient,mansour2020three,Sattler21ClusterFL}. Different from the previous works, this paper aims to improve the algorithm robustness in the presence of both data heterogeneity and device unavailability.
%\vspace{-3ex}
%The CFL aims at dividing clients with similar data distribution into the same cluster, and training a shared model for each client cluster individually. 
%Another approach improves the performance by modifying the aggregation scheme (Bayesian scheme, contribution/similarity weighted).
\subsection{Tackling System Heterogeneity in FL}
Another performance bottleneck faced by FL is the heterogeneous computing capacity and network bandwidth of clients. The straggler issue is caused by the delayed or lost uploading due to system heterogeneity. Existing solutions choose to perform partial client participation \cite{mcmahan17FL} or allow clients to upload their local models asynchronously to the server \cite{ChenNSR20AsynFL,Wu21SAFA, GuHZH21FLunavail,ZhuLLLH21DelayAvg}. A straggler-resilient FL is proposed in \cite{Reisizadeh22StraggFL} that incorporates statistical characteristics of the local training data to actively select clients. An adaptive client sampling scheme is proposed in \cite{LuoXWHT22AdaSampling} to simultaneously tackle the system and statistical heterogeneity. Another line of work aims at dynamically adjusting the local workload according to the device capabilities by changing local batch size and local epochs \cite{Ma21AdaBS,Li21FedSAE}. Client dropout is a more serious problem than stragglers since dropped clients simply have nothing to upload in the current round and can remain unavailable for a relatively long period. Hence, existing solutions for stragglers do not work well for this special issue. Some studies propose re-weighting aggregation rules or variance reduction methods to compensate the missing updates \cite{Jhunjhunwala22Fedvarp,Wang22FDMS}.
\vspace{-2ex}
%In particular, a deadline-based aggregation mechanism could be adopted to address both the straggler and dropout issues, in which aggregation is made once a fixed deadline is met.

\section{Preliminary and Motivation}
\label{sec:Prelim} 

%In this section, we overview the federated learning architecture and non-IID dataset. 

\subsection{Problem Formulation for Federated Learning}
%Federated learning enables clients to learn the shared model collaboratively with two highlights: (i) The decentralized datasets reside on local nodes without being transmitted to the central server; and (ii) Local learners train the model using their local data and only exchange weight parameters through the server. 

A typical implementation of FL system includes a cloud server that maintains a global model and multiple participating clients that communicate with the server through a network. We assume that there are $m$ clients that collectively train a global model under the coordination of cloud server without uploading private data. Mathematically, the learning task could be formalized as the following optimization problem:
\begin{equation}\label{eq:objective}
	\min_{\bm{w}}{L({\bm{w}})} = \sum_{i=1}^{m}\frac{|\mathcal{D}_i|}{|\mathcal{D}|} \underbrace{ \mathbb{E}_{\xi_i \sim \mathcal{D}_i}\left[L_i(\bm{w};\xi_i)\right]}_{:= L_i(\bm{w})},
\end{equation}
where $\mathcal{D}_i$ is the local data set following an underlying distribution $\mathcal{P}_i(X,Y)$ on $\mathcal{X} \times \mathcal{Y}$, where $\mathcal{X}$ is the input space and  $\mathcal{Y} = \{1,.., C\}$ is the label space. $\mathcal{D}$ denotes the collection of all local training data. It is worth noting that generally we have $\mathcal{P}_i \neq \mathcal{P}_j$ for different client $i$ and $j$ in FL. $L_i(\bm{w}_i;\xi_i)$ denotes the local risk given model parameter $\bm{w}$ and data point $\xi_i=(x_i,y_i)$, where $L_i(\cdot;\cdot)$ could be a general objective function, e.g., cross-entropy loss for classification tasks. The global objective $F(\bm{w})$ can be regarded as a weighted average of local objectives and FL tries to minimize it in a distributed manner. For multi-class classification tasks, the local empirical risk is usually the cross-entropy between softmax outputs and ground-truth labels as in Eq.~\eqref{softmax}
\begin{equation}\label{softmax}
	L_i = -\frac{1}{n_i}\sum_{j=1}^{n_i} \sum_{c=1}^{C}\mathbb{I}\{y_j = k\}\cdot \log{\left(\frac{e^{z_{j}[c]}}{\sum_{c'=1}^{C}e^{z_{j}[c']}}\right)},
\end{equation}
where $n_i$ is the local data size, $\mathbb{I}(\cdot)$ is the indication function and $z_{j}$ is the output logit.

The most popular learning method in FL is obviously the FedAvg algorithm~\cite{mcmahan17FL}, which is a strong baseline in many tasks and works as follows. At each communication round $t$, the server selects a random subset $S_t$ of clients to participate in this round of training. The server broadcasts the latest global model ${\bm{w}}^{(t)}$ to these clients for further local optimization. Then each participating client optimizes the local objective function based on the private data by its local solvers (e.g. SGD) with several local epochs $E$ in parallel. Then, the locally-computed parameter updates $\Delta \bm{w}_i^{(t)}$ are collected and aggregated. Finally, the server updates the global model to $\bm{w}^{(t+1)}$ as in eq.~\eqref{agg} and finish the current round. In general, the FL task requires hundreds of rounds to reach the target accuracy. 
\begin{equation}\label{agg}
{\bm{w}}^{(t+1)} = {\bm{w}}^{(t)} + \sum_{i \in S_t} \frac{n_i}{\sum_{i \in S_t} n_i}\Delta{\bm{w}}_{i}^{(t)}
\end{equation}

\subsection{Prior-Corrected Bayes Classifier}

As introduced before, the \textit{Label shift} induced non-IID data in FL refers to the label distribution changing between local training data and target test set. And we focus on the case where local training distributions have significant discrepancy to the uniform test distribution. We start by focusing on the calibration of a single classifier trained on one specific client. Let ${P}_s(X,Y)$ and ${P}_t(X,Y)$ denote the training (source) and the test (target) distributions, respectively. In a multi-class classification task, the final decision on the target distribution is often made by following the Bayes decision rule:
\begin{align}
	\label{eq:bayes_decision}
	y^* = \arg\max_{y} P_t(y|x) &= \arg\max_{y \in \mathcal{Y}} \frac{f_t(x|y)P_t(y)}{P_t(x)} \notag \\
	&= \arg\max_{y\in \mathcal{Y}} f_t(x|y)P_t(y),
\end{align}
where $f_t(x|y)$ is the learned class conditional probability (likelihood) and $P_t(y)$ is the marginal label distribution (prior). Suppose that a discriminative model (e.g., softmax classifier) can perfectly learn the posterior distribution $P_s(Y|X)$ of the source dataset, then the corresponding {Bayes classifier} is defined as the following:
\begin{align}
	h_s(x) = \arg\max_{y\in \mathcal{Y}} P_s(Y=y|X=x).
\end{align}
The classifier $h_s(x)$ is the optimal classifier on the training distribution. By further assuming that the test distribution has the same class conditional likelihood as the training set, i.e. $f_t(X|Y) = f_s(X|Y)$ and differs only on the class priors $P_t(Y) \neq P_s(Y)$. Then the optimal {Bayes classifier} on target distribution is given by the following expression:
\begin{align}
	h_t(x) = \arg\max_{y\in \mathcal{Y}} \frac{P_s(y|x)P_t(y)}{P_s(y)}.
\end{align}

However, in the federated setting, samples on some labels might be totally missing for a local client, which would make the calibration ill-conditioned. One the other hand, notice that
\begin{equation}
	\small
	h_t(x) =\arg\max_{y\in \mathcal{Y}} \frac{P_s(x,y)P_t(y)}{P_s(x)P_s(y)}=\arg\max_{y\in \mathcal{Y}} \frac{P_s(x|y)P_t(y)}{P_s(x)}.
\end{equation}
Therefore, we can alternatively learn the class conditional likelihood $f_s(X|Y)$ on the training data by the technique of \textit{Balanced-SoftMax}, which has been widely applied on long-tailed classification tasks \cite{Ren20BMS,Tian20Posterior}. Specifically, for a data point $(x_j,y_j)$ at client $i$, the log-loss is calculated by
\begin{equation}
	\ell_i(x_j,y_j) = -\log{\left(\frac{p_i(y_j)e^{z_{j}[y_j]}}{\sum_{c=1}^{C}p_i(c)e^{z_{j}[c]}}\right)},
\end{equation}
where $p_i(y)$ is the local prior of label $y$ that can be estimated from the training data. Slightly different from the long-tailed classification, local prior probabilities of some classes could be zero in non-IID FL. Then, the inference on uniform distribution could be performed as the ordinary softmax classifier.

\begin{figure}[t]
	\begin{center}
		%\hspace{1ex}
		\includegraphics[width=0.8\columnwidth,clip=true]{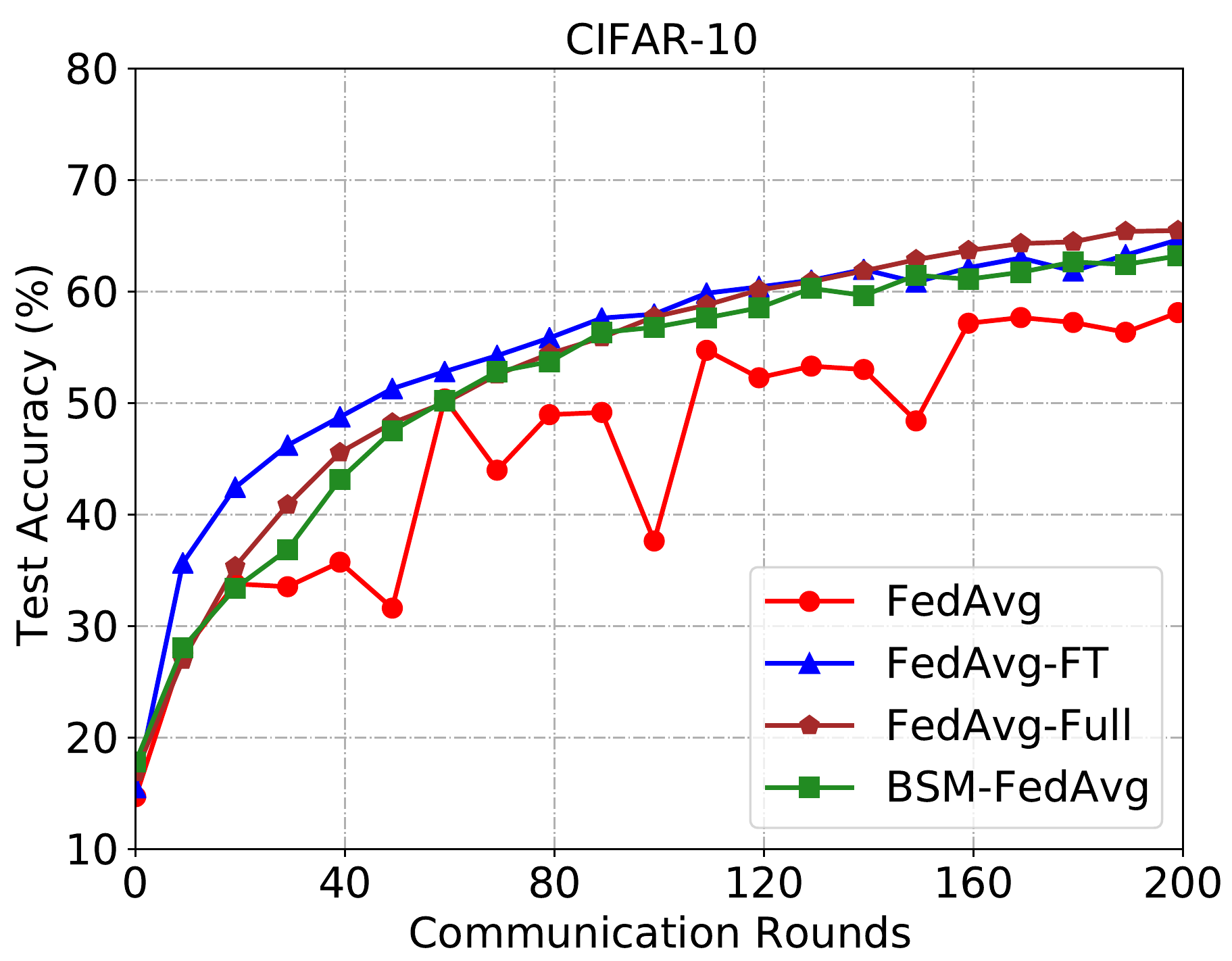}
	\end{center}
	\vspace{-3ex}
	\caption{Preliminary experimental results on CIFAR-10 dataset.}
	\label{fig:cifar_acc_prelim}
	%	\vspace{-2ex}
\end{figure}

\begin{figure*}[ht]	
	{	
		\begin{center}
			%\hspace{1ex}
			\includegraphics[width=1.96\columnwidth,clip=true]{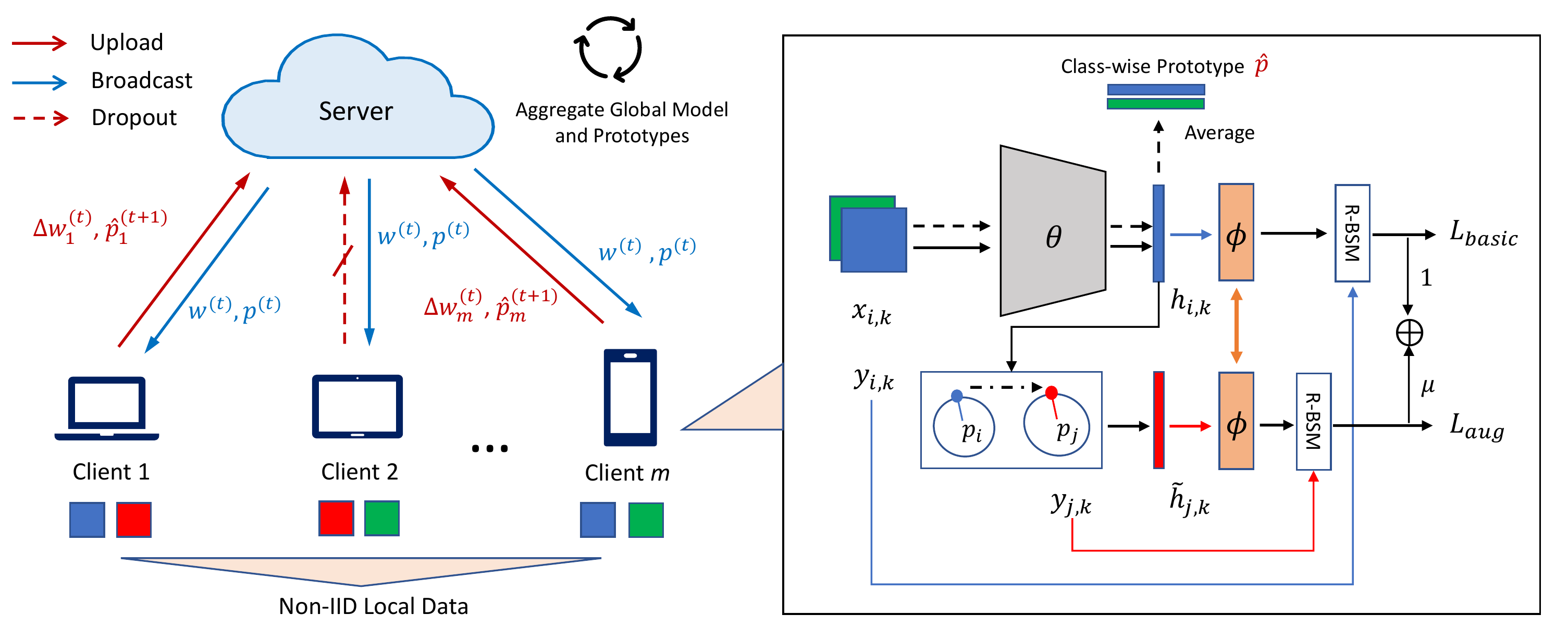}
		\end{center}
	}
	\vspace{-1ex}
	\caption{Federated learning with a central server and a number of clients, where the clients may dropout during training. The local training consists of two key components, including relaxed balanced-softmax and feature-level augmentation by inter-class transfer.}
	%	\vspace{-2ex}
	\label{fig:system}
\end{figure*}

\subsection{Preliminary Results}
We conduct some preliminary experiments on CIFAR-10 to show that the unstable performance of FedAvg on non-IID data is mainly caused by the biased classifier head and the learned representation layers are still effective. To validate this, we assume that a small balanced set of samples is available at the server side for fine-tuning the classifier head in each round. We consider a setup with 50 clients, where each client has samples from only 2 classes and the successful transmission probability is 0.2 in each round. Three methods including vanilla FedAvg, FedAvg with balanced-softmax (BSM-FedAvg) and FedAvg with server-side fine-tuning (FedAvg-FT) are compared with learning curves plotted in Fig~\ref{fig:cifar_acc_prelim}, where we also provide the result of FedAvg with full participation as a reference. The experimental results show that FedAvg with non-IID data and client dropout can result in severe performance fluctuating, while re-training the classifier on a small balanced set can significantly improve the stability and accuracy. On the other hand, it is interesting to see that BSM-FedAvg without the fine-tuning can almost approach the effect of head re-training. This preliminary results motivate us to learn a balanced classifier head without severe forgetting during the local training.

%\subsection{Cluster Structure in FL}
%
%Clusterability of HAR Data
%Although the non-IID data distribution and unstable clients of large amount remain challenging in FL systems, it is useful to notice that the clients in quite a few real-word systems tend to be clusterable in terms of data distribution. For example, in an Internet of vehicles system, vehicles within a certain area tend to record similar transportation information. Besides, the devices within the same smart home system usually collect the features of the same person. In these examples, the dissimilarity between client data in a certain group may be negligible, or even follow IID data distribution for the same learning task.
%Group-wise PFL, which is also named clustered FL, assumes the clients can be clustered to different groups with severe non-IID across inter-group clients and mild non-IID across intra-group clients. Hence, clustered FL can be categorized according to different clustering methods and distance measurements.

\section{Our Proposed Framework}
\label{sec:Framework} 

In this section, we present our framework for FL with non-IID data and client dropout as in Fig.~\ref{fig:system} by introducing two bias-reduction methods in local training, including the modified cross-entropy computing and feature-level augmentation. It is worth noting that the central server is only responsible for message collection and aggregation without any extra model training or other complicated computation.

\subsection{Relaxed Balanced-SoftMax Classifier}

Although the balanced-softmax is effective in tackling the non-IID data to an extent, we claim that a combination of balanced-softmax and vanilla softmax could exhibit better performance in the case of pathological data setting where each client can only observe a few number of classes. As a conceptual example, consider an extreme case where each client only has samples from one class, then the balanced-softmax will result in zero loss and the local training will get stuck, while vanilla FedAvg could still update the model. Moreover, the classifier parameters corresponding to the missing classes could be too stale under the scheme of balanced-softmax, which might be sub-optimal for global aggregation and reduce the overall model performance. Therefore, we propose to use a linear combination of vanilla softmax and balanced-softmax, which could be achieved by the following relaxed version of balanced-softmax (R-BSM):

\begin{equation}
\ell_i(x_j,y_j;\epsilon_i) = -\log{\left(\frac{[\hat{p}_i(y_j)] e^{z_{j}[y_j]}}{\sum_{c=1}^{C}[\hat{p}_i(c)]e^{z_{j}[c]}}\right)},
\end{equation}
and the smoothed prior ${\hat{p}_i}$ is obtained by
\begin{equation}
	\hat{p}_i(y_j) = (1-\epsilon_i) \cdot \frac{n_{i,j}}{n_i} + \epsilon_i \cdot \frac{1}{C},
\end{equation}
where $\epsilon_i \in [0,1]$ is a positive and adjustable coefficient (usually small) added for client $i$. It is interesting to see that this form is very similar to the technique of label smoothing. However, employing label smoothing in local training of FL is not effective as we found in empirical evaluation.

To further analyze the effect of relaxed balanced-softmax, we represent the classifier parameters $\bm{\phi}_i$ of the client $i$ as the weight vectors $\{\bm{\phi}_{i,c} \}_{c=1}^C$, where $\bm{\phi}_{i,c} $ is named the $c$-th proxy as in \cite{Li21FedRS}. For simplicity, we set all the bias vectors of the classifier as zeros and use the cross-entropy loss. Let $q_{j, c} $ denote the normalized score of class $c$ for data point $(\mathbf{x}_j,y_j)$ by the relaxed balanced-softmax, which is written by
\begin{equation}
q_{j, c} =  \frac{\hat{p}_i(c) e^{\bm{\phi}_{i,c}^\mathsf{T} \mathbf{h}_{y_j}}}{\sum_{c'=1}^{C} \hat{p}_i(c') e^{\bm{\phi}_{i,c'}^\mathsf{T}. \mathbf{h}_{y_j}}} 
\end{equation}
Then, the local objective function can be represented as:
\begin{equation}
\begin{aligned}
\mathcal{L}_{i} =-\frac{1}{n_i}\sum_{j=1}^{n_i} \sum_{c=1}^{C} \mathbb{I}\left\{y_{j}=c\right\} \log \left(q_{j, c} \right),
\end{aligned}
\label{sup loss}
\end{equation}
where $\mathbf{h}_{i,y_j} = f_{\mathbf{\theta}}(\mathbf{x}_j)$ is the feature embedding and $\mathbf{\theta}$ denotes the feature extractor. The benefits of applying relaxed balanced-softmax are illustrated as follows.

{{Firstly}, the smoothed prior can mitigate the staleness of proxies}. For the $c$-th proxy $\bm{\phi}_{i,c}$ of the $i$-th client, the
\textit{positive features} and \textit{negative features} denote the features from the $c$-th class and other classes, respectively. The one-step update of $ \bm{\phi}_{i,c}$ with the learning rate $\eta$ can be computed as:
\begin{equation}
\begin{aligned}
\Delta \bm{\phi}_{i,c} = \underbrace{\frac{\eta}{n_i}\sum_{j=1, y_{j}=c}^{n_i}\left(1-q_{j, c}\right) \mathbf{h}_{j}}_{\text {pulling by \textit{positive features}}}
- \underbrace{\frac{\eta}{n_i} \sum_{j=1, y_{j} \neq c}^{n_i} q_{j, c} \mathbf{h}_{j}}_{\text {pushing by \textit{negative features}}}.
\end{aligned}
\label{gradient of classifier}
\end{equation}
For class $c \in \{\mathcal{Y} \setminus \mathcal{Y}_i \}$, there is no update of $\bm{\phi}_{i,c}$ under the original balanced-softmax scheme since the client does not hold any sample of class $c$ and thus $p_i(c)=0$. Therefore, not only there is no pulling force from positive features but also no pushing force from negative features exists due to $q_{j,c} \propto p_i(c)$. Under the scheme of vanilla softmax, the $c$-th proxy is only pushed away from negative features without any restriction from positive features and will result in inaccurate updating (catastrophic forgetting). By contrast, our relaxed version of balanced-softmax sets the prior of any missing class $c$ a small value and thus restricts the pushing force from negative features, striking a balance between the staleness and inaccurate updating of the $c$-th proxy.

%That is, the updates of the $c$-th proxy (i.e., $ \Delta \bm{\phi}_{i,c}$) only depend on \textit{negative features} such that clients that hold different $\{[C]\setminus[C_i]\}$ update the $c$-th proxy inconsistently.
%(ii) for  $c \in [C_i]$, due to  $\mathbf{h}_{i,c}=\mathbf{h}_{j,c}$, the inconsistent $ \Delta \bm{\phi}_{i,c}$ originates from the different updates by \textit{negative features}. This is because the updates depend on the non-$c$-th classes in $[C_i]$, and clients with the skew of $[C_i]$  hold different $\{[C_i]\setminus c\}$.
{{Secondly}, the balanced classifier could mitigate inaccurate feature updating}. Similarly, the update of $\mathbf{h}_{j}$ is represented by:
\begin{equation}
\begin{aligned}
\Delta \mathbf{h}_{j} & = \underbrace{\frac{\eta}{n_i}\left(1-q_{j, y_{j}}\right) \mathbf{\bm {\phi}}_{i,y_{j}}}_{\text {pulling by \textit{positive proxy}  }}
-\frac{\eta}{n_i}\underbrace{\sum_{c \in \mathcal{Y}_i,c\neq y_j} q_{j, c} \mathbf{\bm {\phi}}_{i,c}}_{ \text {pushing by \textit{negative proxies (I)}}}\\
&-\frac{\eta}{n_i}\underbrace{\sum_{c \in {\mathcal{Y} \setminus \mathcal{Y}_i},c\neq y_j} q_{j, c} \mathbf{\bm {\phi}}_{i,c}}_{ \text {pushing by \textit{negative proxies (II)}}}.
\end{aligned}
\label{gradient of feature}
\end{equation}
Similar to the above analysis of proxy updates, the pushing force from any missing class $c$ for updating feature embedding, i.e., the third term in Eq.~\eqref{gradient of feature}, would disappear under both the vanilla softmax and balanced-softmax schemes, leading to inaccurate updates. Specifically, the balanced-softmax directly sets $q_{j,c}$ to zero while the vanilla softmax would result in significantly inaccurate proxy $\mathbf{\bm {\phi}}_{i,c}$ as local training going on and finally the $q_{j,c}$ would also be extremely small.

{Finally}, as the feature and classifier learning are highly coupled during the local training, mitigating the inaccurate updating of them can benefit each other and thus alleviate the performance degradation of federated learning with label skew.

\subsection{Feature Augmentation for Missing Classes}
\label{sec:augmentation}
Although the balanced-softmax can mitigate the impact of missing classes to an extent, the absence of training samples from those classes is still an issue remaining unsolved. One possible way is to generate synthetic data as augmented training samples. However, generating high-quality samples might be not practical in the resource constrained IoT scenarios as the training of generative model will involve high computation and communication costs. To remedy this, we propose to use a lightweight inter-class feature transfer approach to perform the feature augmentation for missing classes, aiming at training a classifier with better generalization ability. 

Following the previous work in imbalanced learning \cite{Yin19FTL}, we assume that the feature ${\bf{h}}_{i,k}$ from $i$-th class lies in a Gaussian distribution with a class feature center ${\bf{p}}_{i}$, a.k.a, prototype, and a covariance matrix $\bf{\Sigma}_{i}$. The class-wise feature centers are estimated as an average over all feature vectors from the same class by a neural network as the feature extractor. To transfer the intra-class variance from local available classes to missing classes, we assume the covariance matrices are shared across all classes, i.e., ${\bf{\Sigma}}_{i}\,{=}\,\bf{\Sigma}$. Theoretically, the augmented feature sample for $i$-th class could be obtained by sampling a noise vector $\bf{\epsilon}\,{\sim}\,\mathcal{N}({\bf{0}},\bf{\Sigma})$ and adding to its center ${\bf{p}}_{i}$. However, it is hard to accurately estimate the covariance matrix and the direction of noise vector might be too random. Therefore, we directly transfer the intra-class variance evaluated from the samples of available classes to the missing classes by linear translation and scaling. 
%{First, we calculate the covariance matrix $\bf{V}$ via:
%\begin{equation}
%	{\bf{V}} = \sum_{i=1}^{N_c} \sum_{k=1}^{m_i} ({\bf{g}}_{ik} - {\bf{c}}_{i})^{T}({\bf{g}}_{ik} - {\bf{c}}_{i})
%\end{equation}
%where $m_i$ is the total number of samples for class $i$. We perform PCA to decompose $\bf{V}$ into major components and take the first $150$ Eigenvectors as ${\bf{Q}}\in\mathbb{R}^{320\times150}$, which preserves $95\%$ energy.}
Specifically, our center-based feature transfer is achieved via:
\begin{equation}
	\tilde{\bf{h}}_{j,k} = {\bf{p}}_{j} + \lambda({\bf{h}}_{i,k} - {\bf{p}}_{i}),
	\label{eq:linearTransfer}
\end{equation}
where ${\bf{h}}_{i,k}$ and ${\bf{p}}_i$ are the feature-level sample and the center of class $i$. ${\bf{p}}_j$ is the feature center of a missing class $j$ and $\tilde{\bf{h}}_{j,k}$ is the transferred feature for class $j$, preserving the same class identity as ${\bf{p}}_j$. This strategy is advantageous as it can be implemented efficiently without repeatedly computing the covariance matrix. The $\lambda$ is an optional coefficient for scaling and we empirically find that set $\lambda$=1 by default is generally good. By sufficiently sampling ${\bf{h}}_{i,k}$ across different available classes, we expect to obtain an augmented distribution of missing class $j$ in the feature space. A more efficient and effective implementation is augmenting all classes by enumerating the label set in a cyclic way until the mini-batch features are all utilized in each gradient decent step.

\begin{algorithm}[ht] 
	\setstretch{1.05}
	\caption{$\mathtt{ReBaFL}$} 
	\textbf{Input:} Learning rate $\eta$, communication rounds $T$, number of clients $m$, local epochs $E$, hyper-parameters $\epsilon$  and $\mu$ \\
	\textbf{Initialize:} Initial model $\bm{w}_0\in R^d$ \\
	\textbf{Server Executes:}
	\begin{algorithmic}[1]
		
		\For{$t=0, 1, ..., T-1$}
		\State Broadcast ${\bm{w}}^{(t)}$ and $\bm{p}^{(t)}$ to clients
		\For {\textit{each client $i$ in active set $S_t$ in parallel}}
		\vspace{0.5ex}
		\State $\Delta{\bm{w}}_{i}^{(t)}$,  $\hat{\bm{p}}_{i}^{(t+1)}$ $\leftarrow$ \textbf{LocalTraining}($i$, ${\bm{w}}^{(t)}$, $\bm{p}^{(t)}$)
		\vspace{0.5ex}
		\EndFor 
		\State Server aggregates $\{\Delta{\bm{w}}_{i}^{(t)}\}$ and updates model:
		\vspace{0.5ex}
		\State \qquad ${\bm{w}}^{(t+1)} = {\bm{w}}^{(t)} + \sum_{i \in S_t} \frac{n_i}{\sum_{i \in S_t} n_i}\Delta{\bm{w}}_{i}^{(t)}$
		\vspace{0.2ex}
		\State Server aggregates prototypes, for $c \in \mathcal{Y}$:
		\vspace{0.5ex}
		\State \qquad $\bm{p}^{(t+1)}[c] = \sum_{i \in S_t} \frac{n_{i,c}}{\sum_{i \in S_t} n_{i,c}} \hat{\bm{p}}_{i}^{(t+1)}[c]$
		\EndFor 
		\vspace{1ex}
		
	\end{algorithmic} 
	
	\textbf{LocalTraining}($i$, ${\bm{w}}^{(t)}$, $\bm{p}^{(t)}$):
	\begin{algorithmic}[1]
		\For{\textit{local epoch} $e=0, 1, ..., E-1$}
		\For{\textit{mini-batch} $\xi_i \in \mathcal{D}_i$}
		\State Local update: ${\bm{w}}_{i}^{(t)} \leftarrow {\bm{w}}_{i}^{(t)} -\eta \nabla_{\bm{w}}L_i(\bm{w}_{i}^{(t)})$
		\EndFor
		\EndFor
		\State Obtain local feature prototypes: $\hat{\bm{p}}_{i}^{(t+1)}[c]$, $c\in \mathcal{Y}_i$
		\vspace{1ex}
		\State \Return $\Delta{\bm{w}}_{i}^{(t)}:= {\bm{w}}_{i}^{(t)}-{\bm{w}}^{(t)}$ and $\hat{\bm{p}}_{i}^{(t+1)}$ back
		\vspace{1ex}
		
	\end{algorithmic} 
	
	\label{alg1}
\end{algorithm}

\subsection{Algorithm Design}
\vspace{1ex}
\noindent \textbf{Local Optimization Objective.} 
To strike a balance between exploring local knowledge and exploiting global knowledge, a client needs to update the local model by considering both local samples and global prototypes. To this end, we incorporate the designed relaxed softmax and feature-prototype based augmentation to learn a balanced classifier. Then, the local objective function can be written as follows:
\begin{equation}
	\min L_i({\bm w}_{i}):=\frac{1}{n_i} \sum_{j=1}^{n_i} \left[\ell_i({\bm w}_{i};\xi_j)+ {\mu}\cdot \ell_i(\bm{\phi}_i;\tilde{{\bf{h}}}_j) \right],
	\label{eq:loss}
\end{equation}
where $\tilde{{\bf{h}}}_j$ is the augmented feature embedding based on the $j$-th data point and prototypes of class $y_j$ and class $y_{j'}:= [j\mod C]$, $\mu$ is a hyper-parameter to balance the two terms of loss. It is worth noting that the global prototypes might not contain all classes, therefore we only augment those available classes and re-calculate the prior of augmented features for computing the cross-entropy loss. The main workflow of ReBaFL is presented in Algorithm~\ref{alg1}, where the optimization is implemented in an iterative manner with multiple client-server communication rounds.

\vspace{0.5ex}
\noindent \textbf{Local Training Procedure.} 
After receiving the latest global model, the clients first update the model parameters: 
\begin{equation}
	{\bm w}_{i}^{(t)} \leftarrow {\bm w}^{(t)}.
	\label{eq:local_initial}
\end{equation}
Then, the class-wise prototypes are obtained by combining the received global prototypes ${\bm{p}}^{(t)}$ and the newly computed local prototypes ${\bm{p}}_{i}^{(t)}$ by the received feature extractor:
\begin{equation}
\bm{p}_c^{(t)} \leftarrow {\bm{p}}_{i}^{(t)}[c], \quad \forall c \in \mathcal{Y}_i.
\label{eq:new_proto}
\end{equation}
This step is designed mainly in consideration of the possible missing prototypes from previous round. Finally, the mini-batch stochastic gradient decent (SGD) method will be applied to update the whole model for multiple epochs:
\begin{equation}
	\nabla_{\bm w}L_i({\bm w}_{i}^{(t)}) = \nabla_{\bm w}\ell_i({\bm w}_{i}^{(t)};\xi_i) + {\mu}\nabla_{{\bm \phi}}\ell_i({\bm \phi}_{i}^{(t)};{\bf{h}}_i),
	\label{eq:local_grad}
\end{equation}
\begin{equation}
	{\bm w}_{i}^{(t)} = {\bm w}_{i}^{(t)} -\eta \nabla_{\bm w}L_i({\bm w}_{i}^{(t)}),
	\label{eq:local_train}
\end{equation}
where the $\xi_i$ denotes mini-batch data instead of a single data point in practical implementations. It can be easily found that the main differences between our method and FedAvg include employing the relaxed balanced-softmax in computing the local cross-entropy loss and class-ordered cyclic feature-level augmentation. 

%This method has several advantages in the sense that it does not modify the training process in clients and can perform more local training epochs given the same computation overhead as other methods, e.g., Ditto.

\vspace{0.5ex}
\noindent \textbf{Local Prototypes Extraction.} After the local model updating, the local feature prototypes for each class can be computed through a single pass on the local set as follows,
\begin{equation}
	\hat{\bm{p}}_{i,c}^{(t+1)} = \frac{\sum_{j=1}^{n_i} \bm{1}(y_j^{(i)}=c)\bm{f}_{\bm{\theta}_i^{(t+1)}}(x_j^{(i)})}{\sum_{j=1}^{n_i}. \bm{1}(y_j^{(i)}=c)}, \quad \forall c \in \mathcal{Y}.
	\label{eq:local_proto}
\end{equation}

\noindent \textbf{{Global Model Aggregation.}}
%In each communication round, the server receives updated models and feature statistics from clients to perform model aggregation and update global centriods.
%Like the common algorithms, the server performs weighted averaging of local updated model parameters to obtain a new global model, with each coefficient determined by the local data size.
As mentioned before, the server is only responsible for information aggregation and broadcasting, therefore, we directly employ the sample size-based weighted average for global model updating as follows. Investigating more advanced and sophisticated aggregation methods can be a future direction.

\begin{equation}
	{\bm w}^{(t+1)} = \sum_{i=1}^{m} {\alpha_{i}}{\bm w}^{(t)}_{i} , \quad \alpha_{i} = \frac{n_i}{\sum_{i=1}^{m}n_i}.
%	\vspace{-0.5ex}
\end{equation}

\noindent \textbf{Update Global Feature Prototypes.} The global prototype $\bm{p}_c$ for each class $c$ is estimated by aggregating the received local prototypes by the following equation:

\begin{equation}\normalsize
	\bm{p}_c^{(t+1)} = \frac{1}{\sum_{i=1}^{m}n_{i,c}}\sum_{i=1}^{m} n_{i,c} \hat{\bm{p}}_{i,c}^{(t+1)}, \quad \forall c \in \mathcal{Y}.
	\label{eq:c}
\end{equation}

When client dropout occurs, the server will not wait for all the $m$ clients at each synchronization round but only collect messages from active clients within a pre-defined time to mitigate the impact of possible stragglers and improve the system efficiency.

\subsection{Practical Consideration}
\noindent \textbf{Efficiency.}
During the local training, the main difference is the feature-space augmentation and is implemented by the linear inter-class transfer, which is computation-efficient. The computation of local class-wise feature prototypes would induce negligible overhead as only one pass of local data is needed. The transmission of prototypes is also communication-efficient as the number of bits is much smaller than the size of exchanged parameters. Therefore, the proposed framework is efficient and practical.

\vspace{0.5ex}
\noindent \textbf{Privacy.}
Similar to the FedAvg, a relatively large mini-batch size and multiple local epochs are applied for local training, which can offer some extent of privacy guarantee. Our empirical verification will also confirm that the feature prototypes will not raise high privacy risks. Before performing aggregation in the server, anonymization mechanism such as random shuffling \cite{Liu0CGY21FLAME,GirgisDDKS21Shuffled} can also be equipped to hidden the statistical information of individual clients.

\section{Evaluation}\label{secExperiment}

\subsection{Experimental Setup}
\noindent \textbf{Datasets and Models.} Two commonly used benchmark image classification tasks are employed to access our proposed framework, including Fashion-MNIST \cite{fmnist} and CIFAR-10 \cite{cifar10/100}. Both datasets contain 10-category objects and can be easily partitioned to simulate the non-IID distributions in FL. Two convolutional neural networks (CNNs) similar to LeNet-5 are constructed as the learning models for the above two datasets. The first CNN model has two convolution-pooling pairs and two fully-connected (FC) layers, where the convolutional layers have 16 and 32 channels, the FC layers have 128 and 10 neurons, respectively. An additional convolutional layer with 64 channels is used for the second CNN model to increase the model capacity, which will be trained on the CIFAR-10 dataset.

\vspace{0.5ex}
\noindent \textbf{Data Allocation.} Like previous works \cite{mcmahan17FL,zhao2018nonIID}, we select a small set of samples from only a few classes as the local training data for each client. We assume each client $i$ only has samples from $N$ different classes. To be more precise, a subset $\mathcal{Y}_i \subseteq \mathcal{Y}$ with $|\mathcal{Y}_i|=N$ and $n_i/N$ instances of each class  in $\mathcal{Y}_i$ will be randomly sampled and allocated to client $i$. In this paper, we set $n_i=1000$ for all experiments. We focus on the pathological setting where each client only has samples from 2 classes, and also investigate another clustered setting where the clients are evenly divided into multiple groups and clients within the same group have the same number of available classes.

\vspace{0.5ex}
\noindent \textbf{Compared Methods.} The following commonly used baselines and start-of-the-art methods are compared: vanilla FedAvg~\cite{mcmahan17FL}, FedProx \cite{Li20FedProx}, Scaffold \cite{Karimireddy20SCAFFOLD}, FedDyn \cite{Acar21FedDyn}, MOON \cite{LiHS21MOON}, FedBABU \cite{OhKY22FedBABU} and FedRS \cite{Li21FedRS}, FedAvg with balanced-softmax (BSM-FedAvg) \cite{Chen22FedRoD} and FedAvg with logit calibration (FedLC) \cite{Zhang22FedLC}. For the baseline methods, we choose the recommended hyper-parameters as in the original papers. For our ReBaFL method, we tune $\epsilon$ over $\{0.001, 0.01, 0.1\}$ and $\mu$ over $\{0.1, 0.5, 1, 5\}$, and find that choosing $\epsilon=0.01$ and $\mu=0.1$ by default performs generally well for all experiments. Notice that we do not focus on tuning the best combination of hyper-parameters in our framework.

\vspace{0.5ex}
\noindent \textbf{Training Settings.} We consider a moderate-sized FL setup with $m=20/50$ clients and employ the SGD as the optimizer with mini-batch size $B=50$ and weight decay 5e-4. The local learning rate is fixed to 0.01/0.02 for the Fashion-MNIST and CIFAR-10 datasets, respectively. For all experiments, the number of local training epochs is set as $E=5$ to reduce the communication overhead. The total number of communication rounds is set to 200, which is enough for convergence in our experiments. At each round, we record the test accuracy of the global model on the balanced test set that has an uniform label distribution.

\begin{table}[t]
	\small
	\centering
	\caption{The comparison of test accuracy (\%) with system sizes m=20/50 and successful transmission probability $p=0.5$. }
	
	%	\vskip 1ex
	\begin{tabular}{ l c c c c}
		\toprule[1pt]
		\multirow{2}{*}{\textbf{Method}}& \multicolumn{2}{c}{Fashion-MNIST} &
		\multicolumn{2}{c}{CIFAR-10} \\
		\cmidrule(lr){2-3}\cmidrule(lr){4-5}
		& {$m=20$} & {$m=50$}
		& {$m=20$} & {$m=50$} \\
		\midrule
		FedAvg
		& 75.27 & 83.60
		& 54.82 & 64.37
		\\
		FedProx
		& 75.36 & 83.55 
		& 54.66 & 64.14 
		\\
		Scaffold
		& 63.48 & 65.62 
		& 20.57 & 44.56
		\\
		FedDyn
		& 77.98 & 85.26
		& 56.41 & {68.51}
		\\
		MOON
		& 65.35 & 80.35
		& 47.28 & 59.08
		\\
		FedRS
		& 72.46 & 79.83 
		& 54.39 & 64.05 
		\\
		FedBABU
		& 74.20 & 82.34 
		& 52.73 & 62.19
		\\
		FedLC
		& 68.75 & 76.98 
		& 53.56 & 65.70
		\\
		\midrule
		BSM-FedAvg
		& 68.64 & 76.99
		& 53.25 & 65.52
		\\
		\textbf{ReBaFL} (ours)
		& \textbf{78.81} & \textbf{85.44} 
		& \textbf{58.49} & \textbf{68.59} 
		\\
		\bottomrule[1pt]
		\label{tab:result_main}
	\end{tabular}
		\vspace{-2ex}
\end{table}

\begin{figure*}[h]
	\centering
	\subfigure{
		\includegraphics[width=0.63\columnwidth]{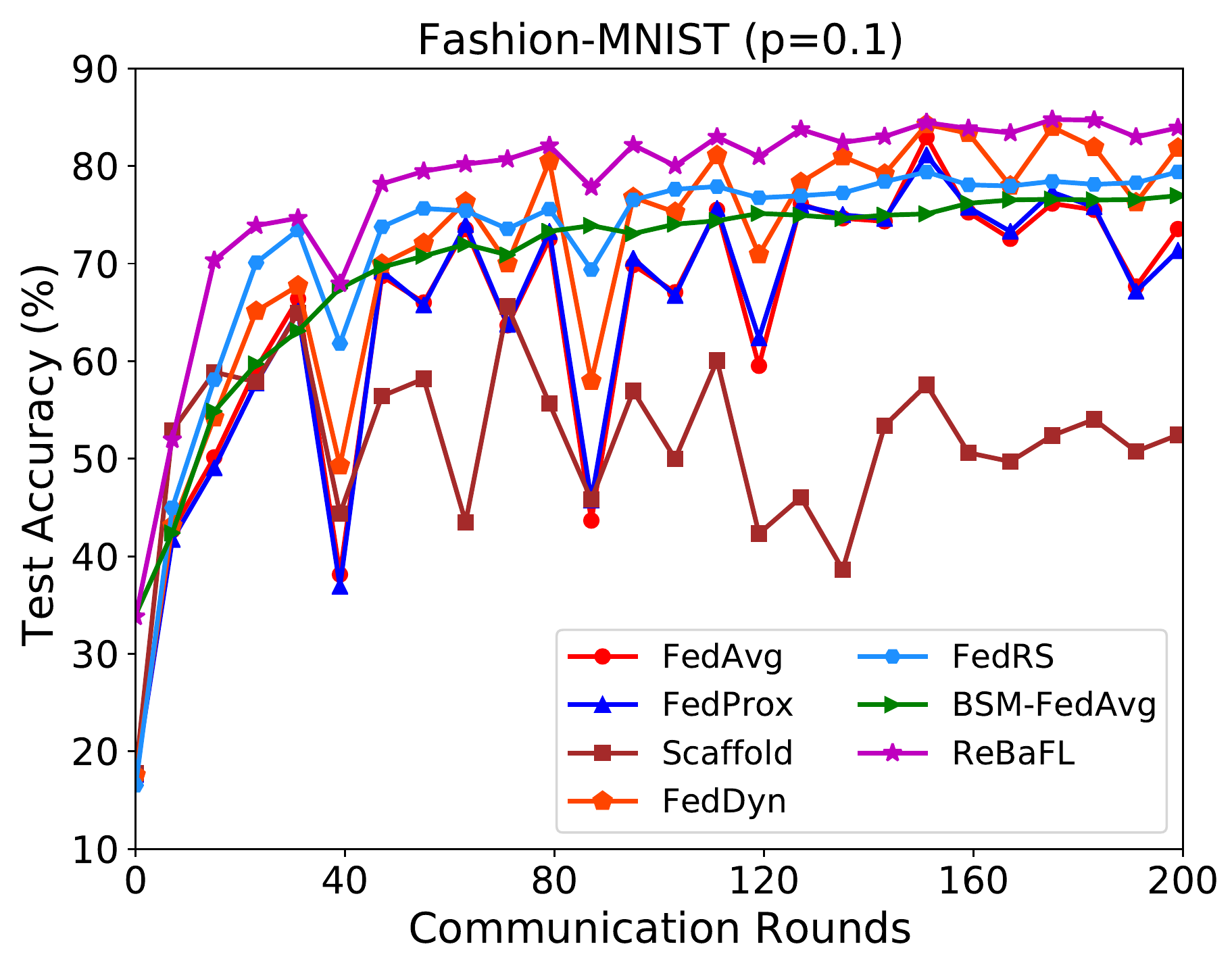}
	}
	%	\quad
	\hspace{-1ex}
	\subfigure{
		\includegraphics[width=0.63\columnwidth]{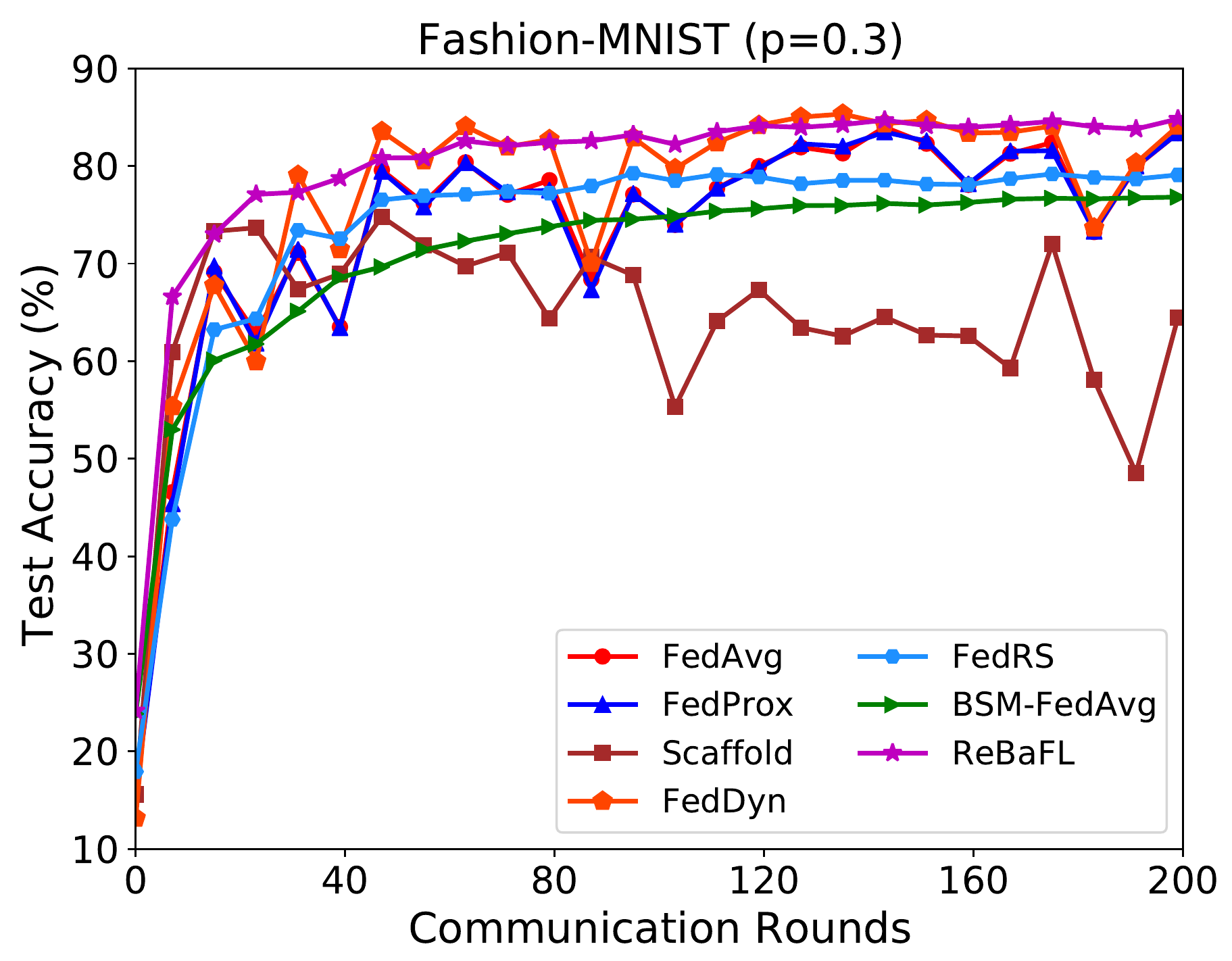}
	}
	%	\quad
	\hspace{-1ex}
	\subfigure{
		\includegraphics[width=0.63\columnwidth]{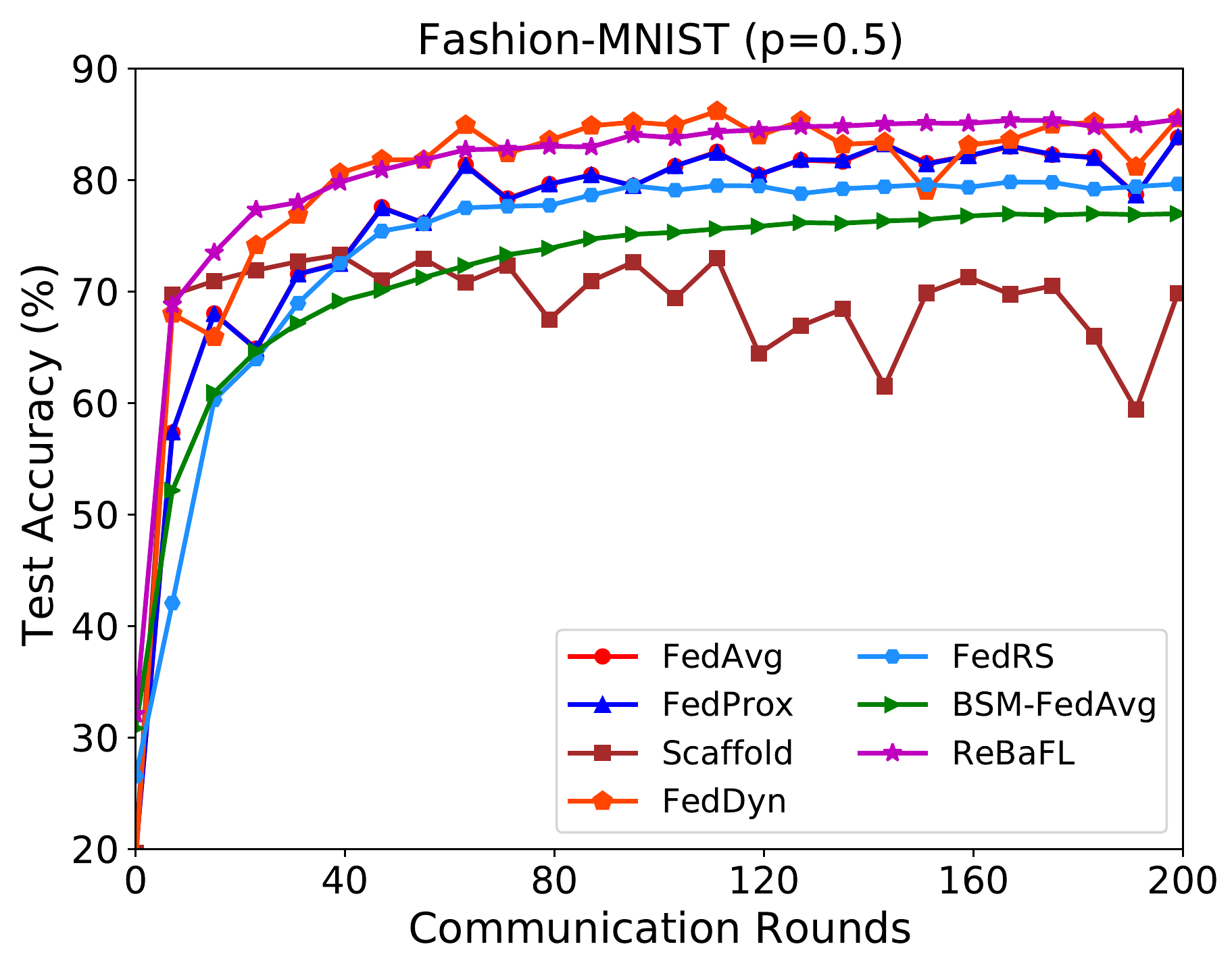}
	}\\
	%	\quad
	%	\vspace{3ex}
	\subfigure{
		\includegraphics[width=0.63\columnwidth]{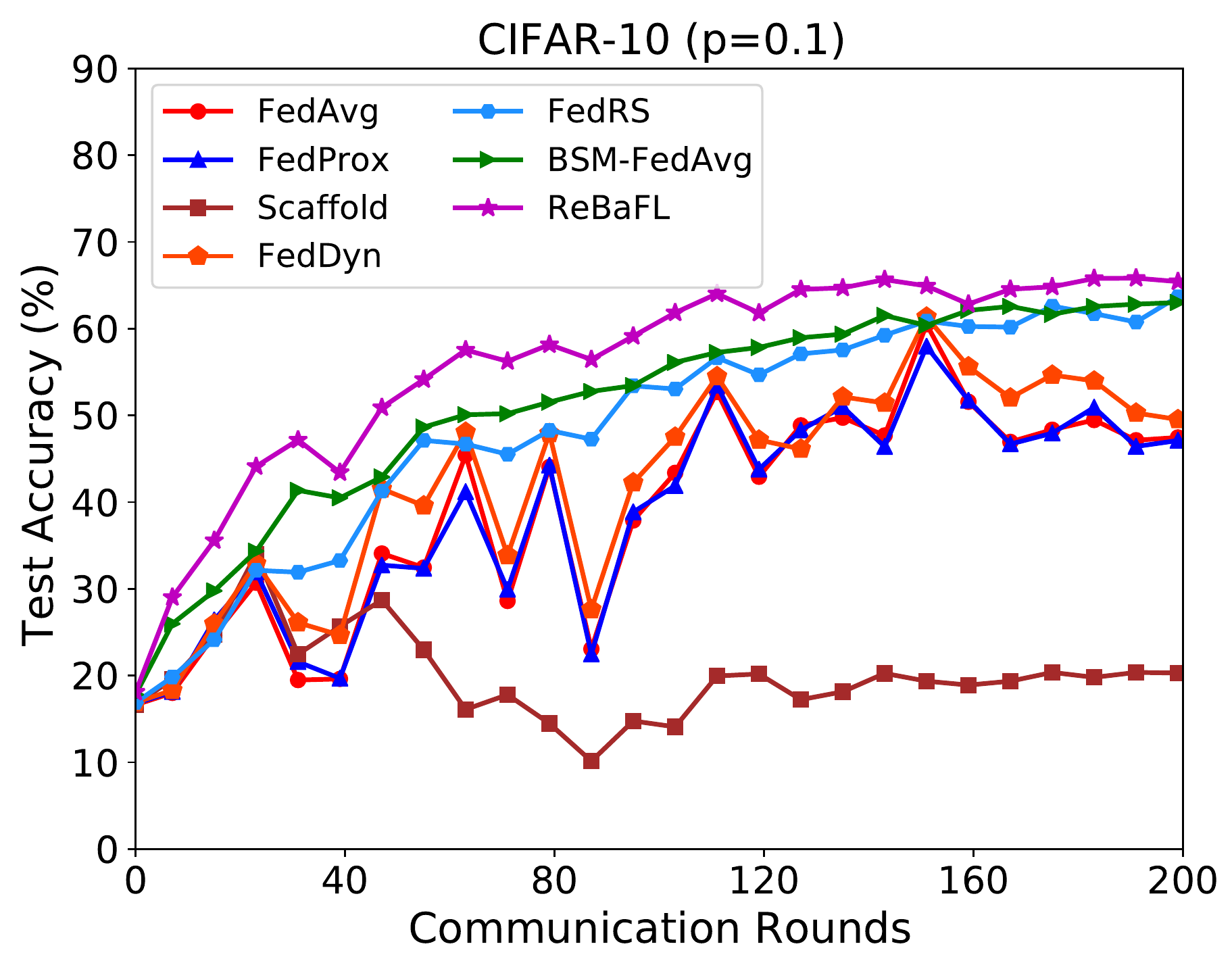}
	}
	%	\quad
	%	\quad
	\hspace{-1ex}
	\subfigure{
		\includegraphics[width=0.63\columnwidth]{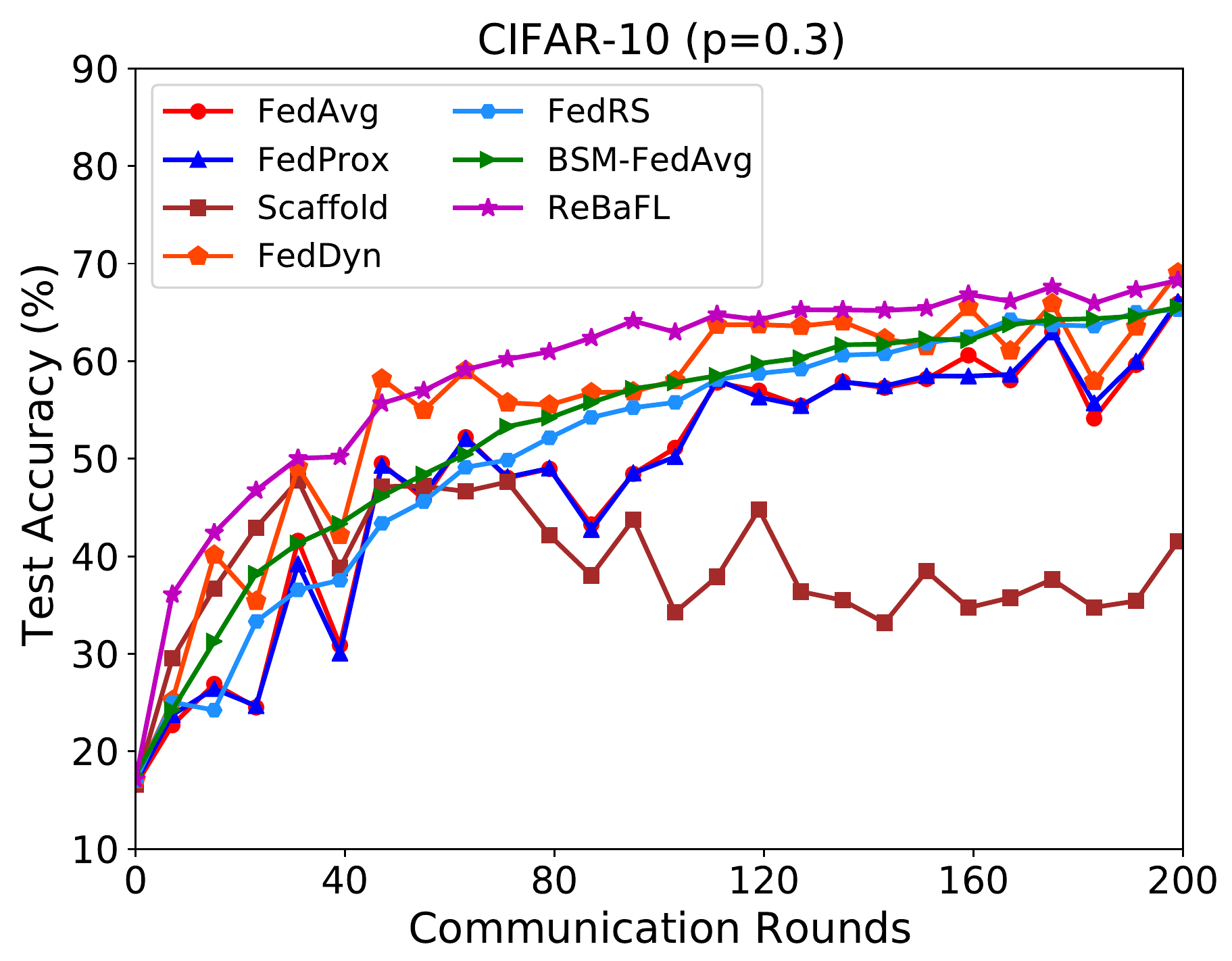}
	}
	%	\quad
	\hspace{-1ex}
	\subfigure{
		\includegraphics[width=0.63\columnwidth]{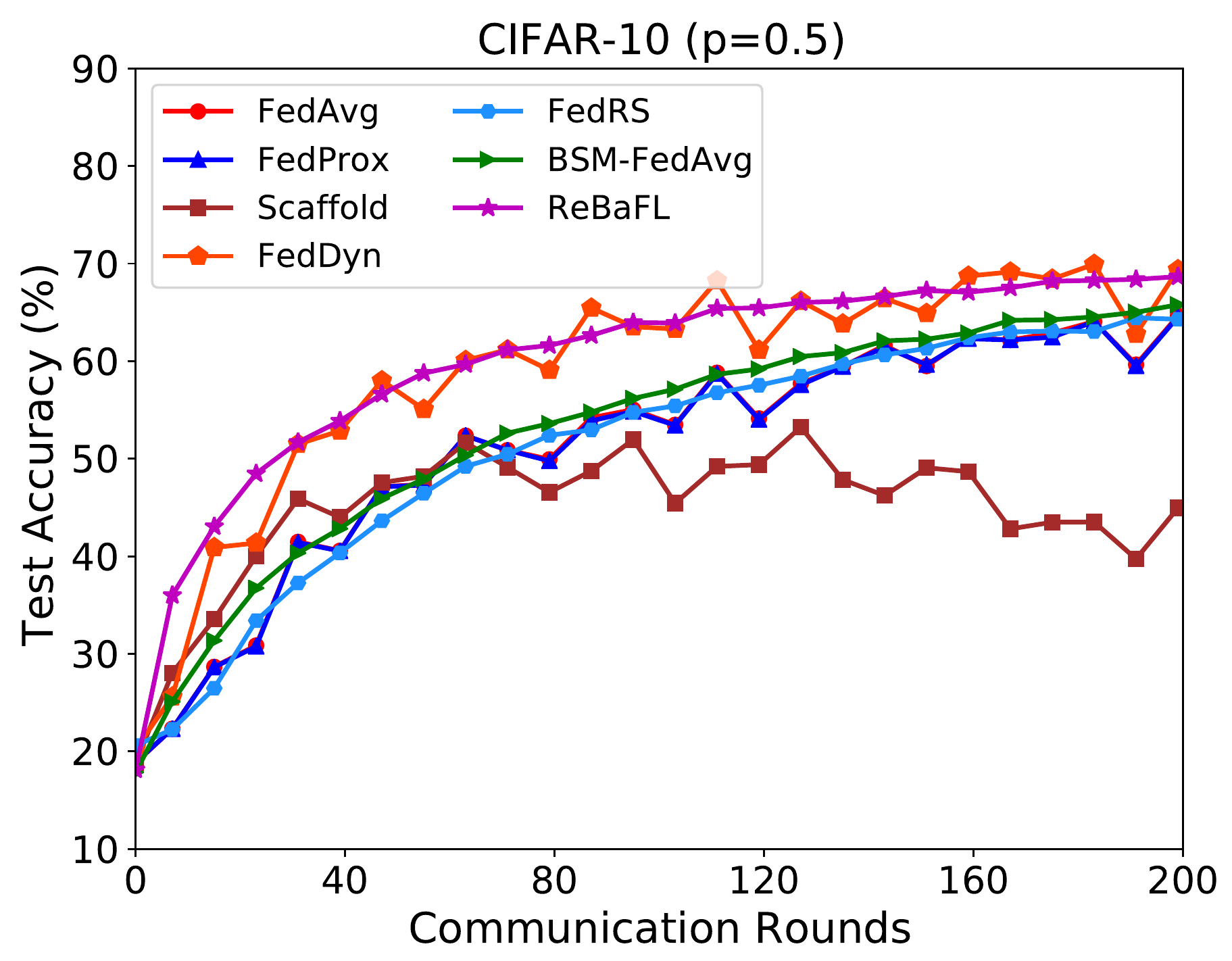}
	}
	%	\quad
	%	\vspace{-1ex}
	\caption{Test accuracy over communication rounds under different levels of client dropout. Each client only has two randomly allocated labels.}
	\label{fig:main_results}
%	\vspace{-2ex}
\end{figure*}

\subsection{Main Results in Pathological Setting}

\noindent \textbf{{Performance Comparison.}} We first study the random and independent straggler scenarios, which can be simulated by applying an unified successful transmission probability $p$ for all clients. All compared methods are evaluated under the same experimental conditions by using the same random seed. From the numerical results in Table~\ref{tab:result_main}, it can be seen that our ReBaFL can consistently outperform other methods in various cases and achieve the highest global test accuracy. The results of FedAvg, FedProx and Scaffold also show that learning on non-IID data with client dropout is particularly difficult due to the unstable training induced by missing classes. In addition, it is also interesting to see that FedAvg is indeed a strong baseline, resulting in comparable or even better performance compared to various advanced algorithms.

\vspace{0.5ex}
\noindent \textbf{{Resilience to Client Dropout.}} To investigate the impact of client dropout, we also vary the $p$ over $\{0.1, 0.3, 0.5\}$ in the setup of 50 clients. The results in Fig.~\ref{fig:main_results} show that the high dropout rate would significantly degrade the performance of FedAvg and other methods that based on the vanilla softmax normalization during local training. Since the low participation rate cannot ensure that samples from all classes are available at each round, the global aggregation would not be able to correct the highly biased local classifier heads. In contrast, our method can stabilize the training performance and exhibit high robustness to various levels of client dropout. Besides, it can also be found that both FedRS and BSM-FedAvg that leverage the restricted softmax can already combat the label skew and client dropout with good training stability. 

The above results indicate that the conventional softmax-based training is vulnerable to the highly-skewed label distribution in FL and the high client dropout rate (i.e., low successful transmission probability) will further worsen the model performance. Instead, applying the balanced-softmax mechanism is usually beneficial and robust to label distribution skew. And our proposed relaxed version of balanced-softmax equipped with feature augmentation can overcome such label skew and client dropout to a larger extent and achieve better model performance.

%\begin{table*}[t]
%	%	\footnotesize
%%	\scriptsize
%    \small
%	\centering
%	\caption{\small Test accuracy (\%). FL system with 20 clients.}
%	\vskip 1ex
%	\renewcommand\arraystretch{1.3}
%	\begin{tabular}{ l c c c c c c c c c}
%		\toprule[1pt]
%		{Dataset}  & {FedAvg} & {FedProx} & {SCAFFOLD} & {FedDyn} & {MOON} &  {FedRS} & {FedBABU} & {BSM-FedAvg} & \textbf{ReBaFL}\\
%		\midrule
%		Fashion-MNIST & 79.48 & 84.03     & 72.77      & 78.63    & 72.60   & 84.12    & 76.65     & 85.08        & \textbf{85.92} \\
%		
%		CIFAR-10      & 59.73 & 66.08     & 31.92      & 57.92    & 56.29   & 64.81    & 57.86     &  66.23       & \textbf{67.12} \\
%		\bottomrule[1pt]
%		\label{tab:cluster_setting}
%	\end{tabular}
%\vspace{-1ex}
%\end{table*}
\begin{figure}[h]
	\begin{center}
		\includegraphics[width=0.95\columnwidth,clip=true]{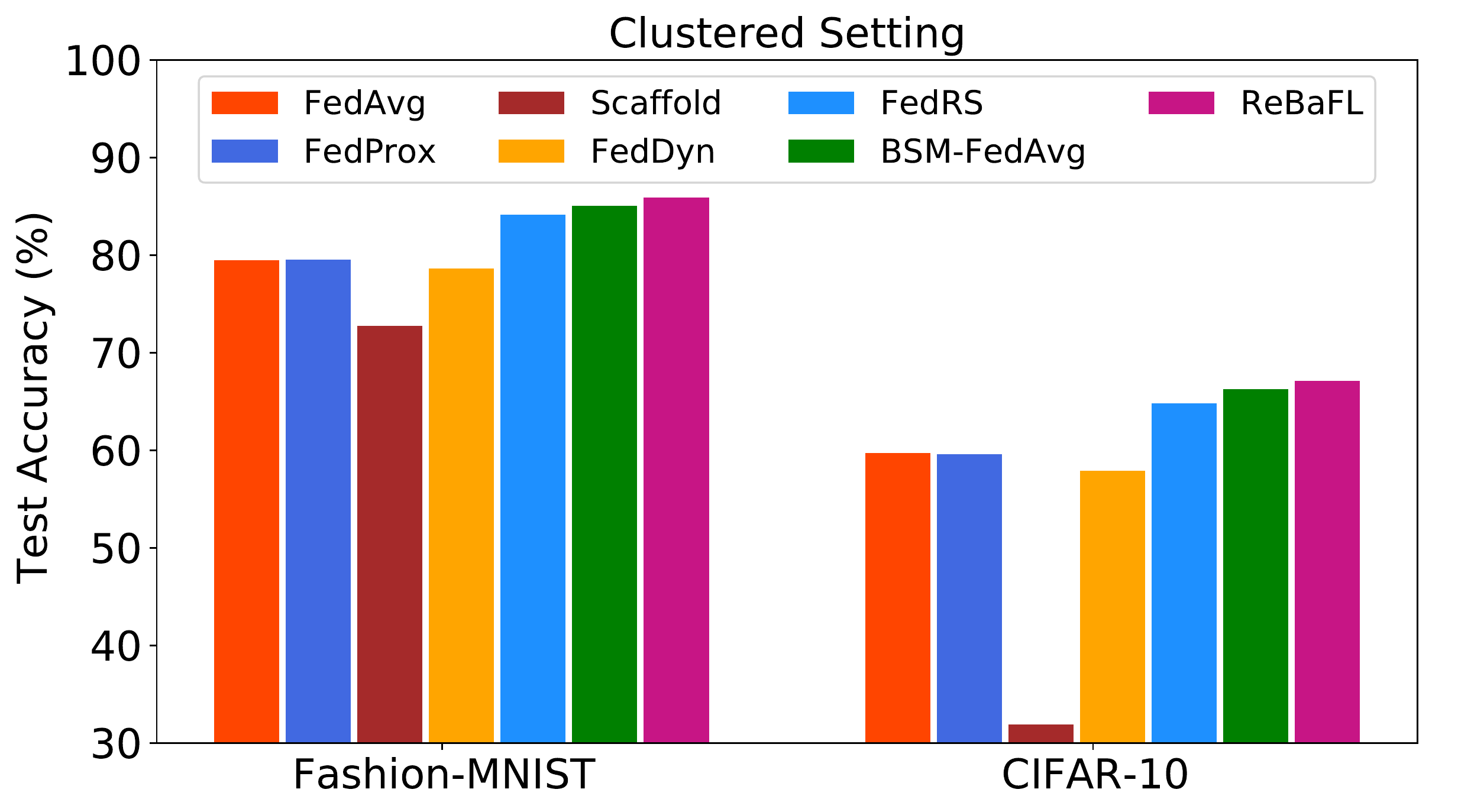}
	\end{center}
	\vspace{-2ex}
	\caption{Performance comparison in clustered setting. Objective calibration-based methods outperform regularization-based methods with large margins.}
%	 where the clients are divided into multiple groups and the intra-group clients share the same number of labels and successful transmission probability
	\label{fig:cluster_bar}
\end{figure}

\begin{figure}[h]
	\vspace{-1ex}
	\centering
	\subfigure{
		\includegraphics[width=0.475\columnwidth]{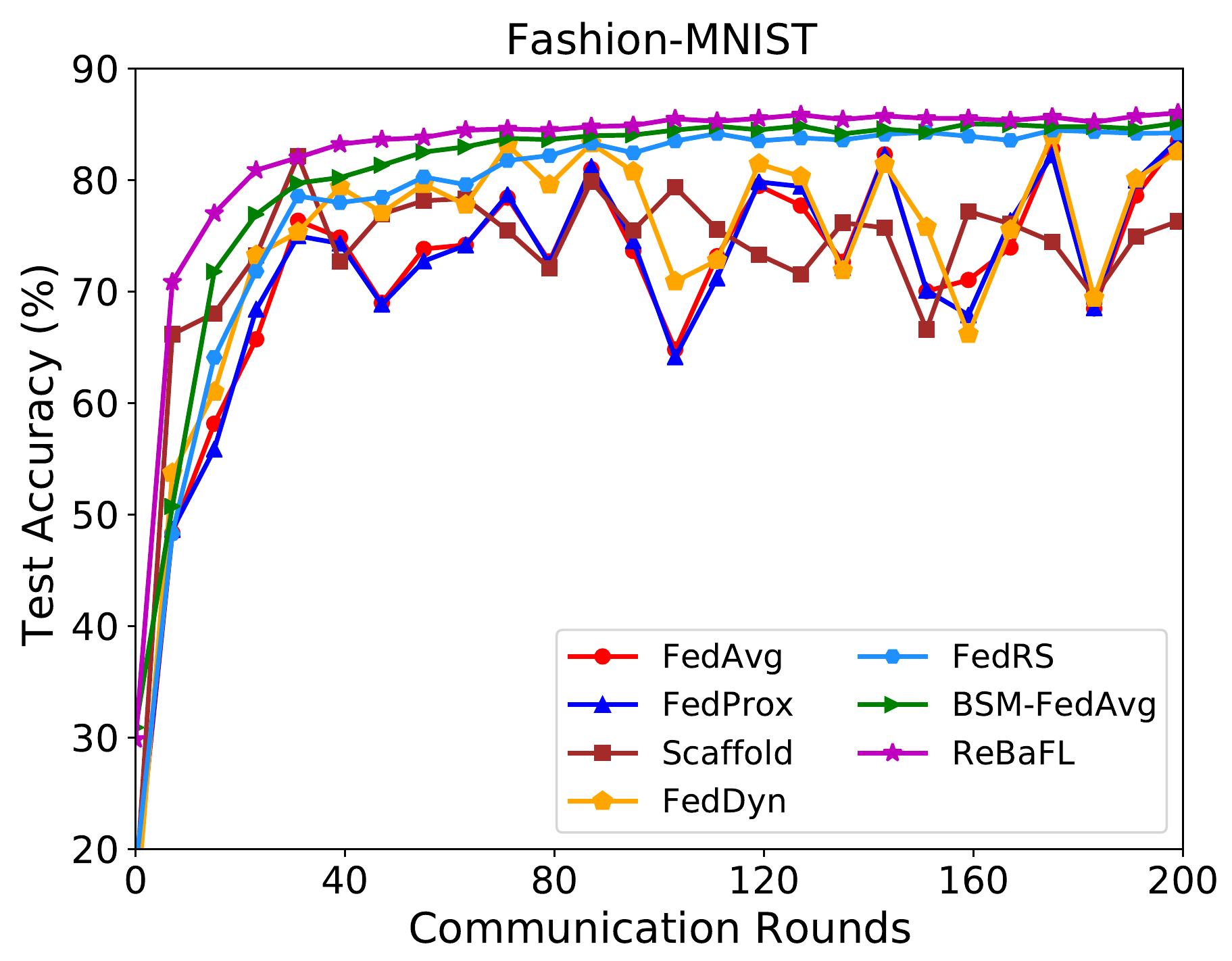}
	}
	%	\quad
	\hspace{-2ex}
	\subfigure{
		\includegraphics[width=0.475\columnwidth]{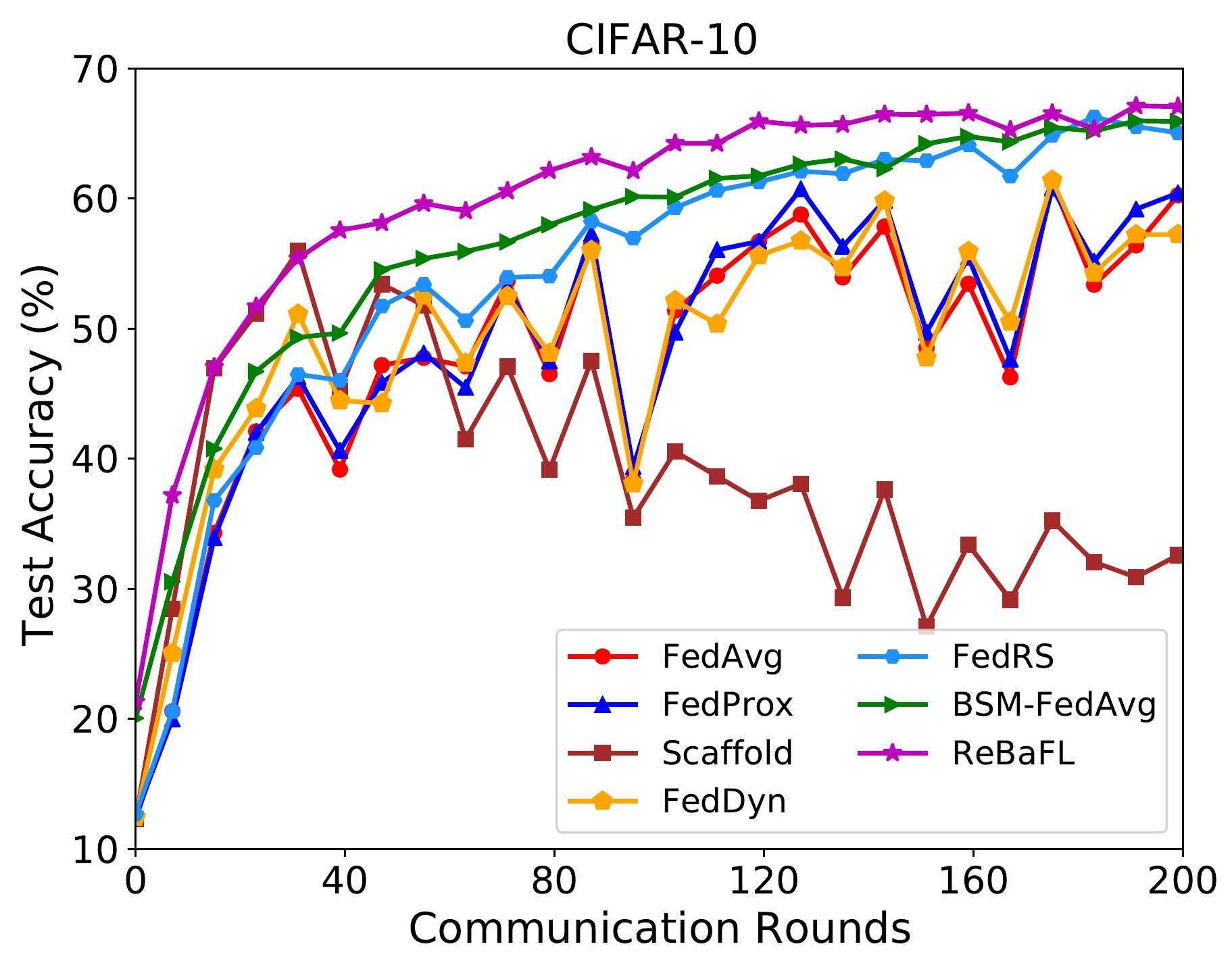}
	}
	%	\quad
	\caption{Test accuracy over communication rounds in clustered setting. ReBaFL can present fast convergence and better model accuracy.}
	\label{fig:main_results_cluster}
		\vspace{-2ex}
\end{figure}

\subsection{Main Results in Clustered Setting}

In this setting, we assume that the clients have clustered structures and consider a setup with 20 clients which are grouped into 5 clusters. In each cluster, the clients share the same number of available classes and the same successful transmission probability, while those quantities could vary across different clusters. In such cases, it is of significant importance to ensure the stability of FL training. We set the number of class $N$ by $\{2,2,3,3,5\}$ and the $p$ by $\{0.5, 0.5, 0.3, 0.3, 0.2\}$ for each cluster, respectively. Notice that even the clients within the same cluster may have substantially distinct subsets of labels. The main results are presented in Fig.~\ref{fig:cluster_bar} and Fig.~\ref{fig:main_results_cluster}, from which we can see that, overall, the ReBaFL still achieves the best model performance and training stability during the federation procedure with client dropout. In this setting, the FedDyn tends to be as unstable as FedAvg due to label distribution skewness and unexpected client dropout. It can also be found that BSM-FedAvg and FedRS also present good performance since the data heterogeneity is relatively mild compared to the pathological setting.

\subsection{Ablation Studies}

\subsubsection{Effect of Individual Component}
There are two important components in ReBaFL, i.e., relaxed balanced-softmax (R-BSM) and feature augmentation (F-Aug). Here we provide some ablation studies to investigate the individual effects of them. The experiments are conducted on CIFAR-10 dataset under both Pathological and Clustered settings. The resulted model accuracy is reported in Table~\ref{tab:ablation}, from which we can find that the relaxed-softmax alone can already provide a significant improvement of the model accuracy and the combination of those two components can lead to the best performance and exceed the baseline with a large margin. One thing interesting we can see is that the inter-class feature transfer based augmentation can even worsen the model performance when the balanced softmax is not applied.

\begin{table}[h]
%	\footnotesize
    \small
	\caption{Model Accuracy (\%) with Different Design Components}
	\centering
	\label{tab:ablation}

		\begin{tabular}{ c  c  c  c }
			\toprule
			\multirow{2}{*}{\textbf{R-BSM}} & 
			\multirow{2}{*}{\textbf{F-Aug}} & 
			\multicolumn{2}{c}{\textbf{Data Settings}}\\
			\cline{3-4} 
			& & Pathological & Clustered\\
			\midrule
		    &             & 54.82 & 59.71 \\
			\Checkmark &  & 56.71 & 65.75 \\
			&  \Checkmark & 54.01 & 58.85 \\
			\midrule
			\Checkmark & \Checkmark  & \textbf{58.49} & \textbf{67.12} \\
			\bottomrule
		\end{tabular}
	\vspace{1ex}
	
\end{table}

\begin{figure}[h]
	\centering
	\subfigure[]{
		\includegraphics[width=0.46\columnwidth]{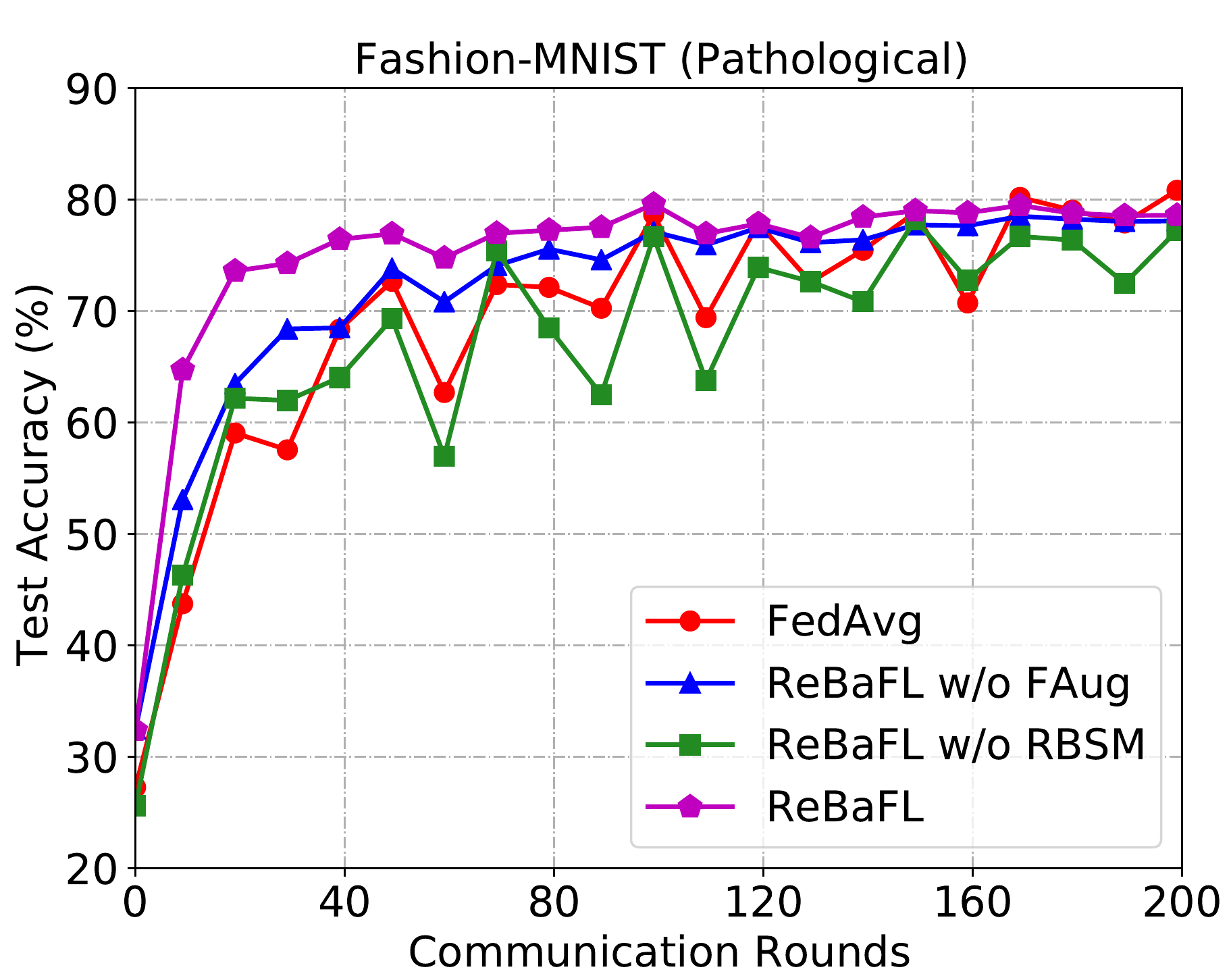}
	}
	%	\quad
	\subfigure[]{
		\includegraphics[width=0.46\columnwidth]{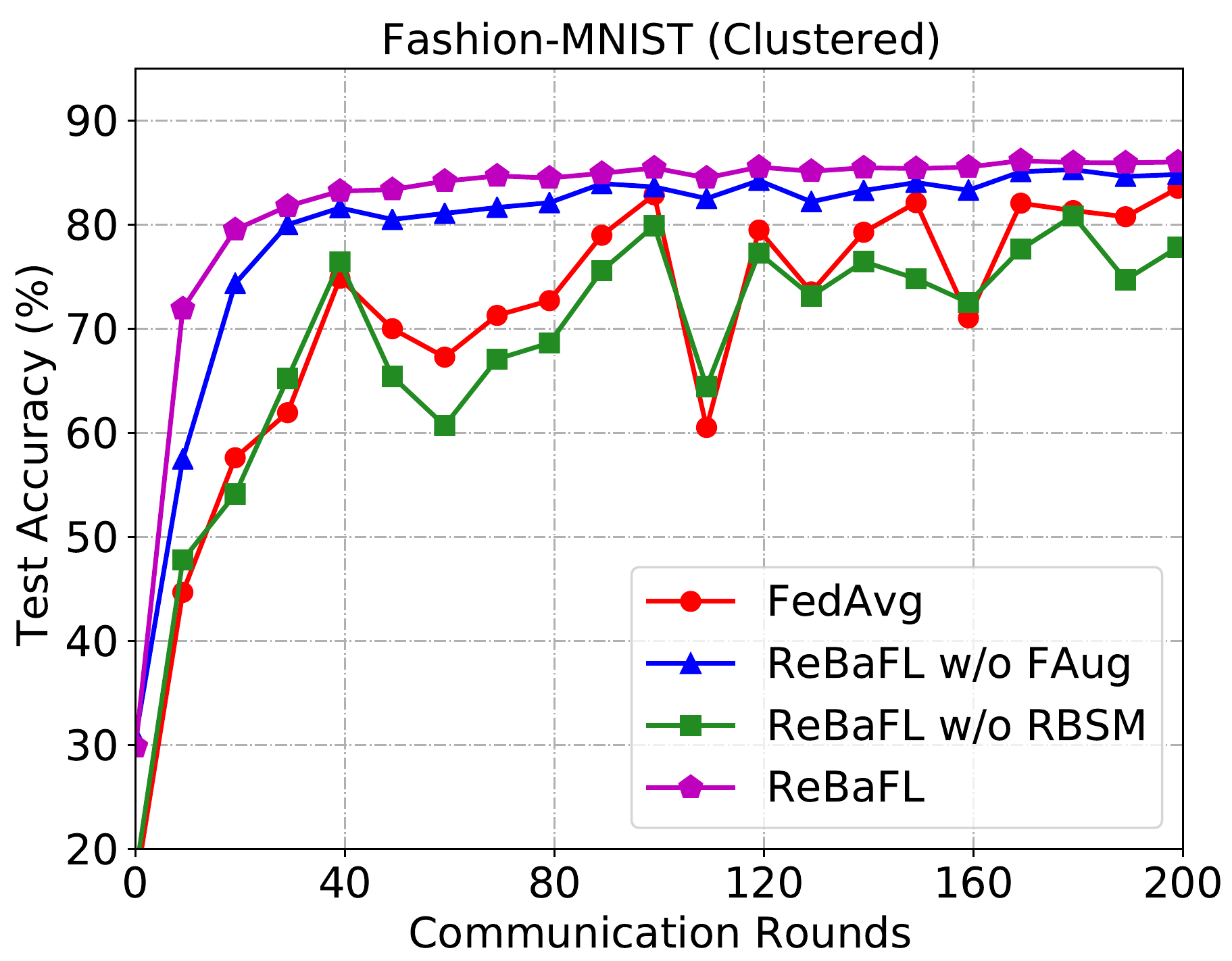}
	}
	%	\quad
	\subfigure[]{
		\includegraphics[width=0.46\columnwidth]{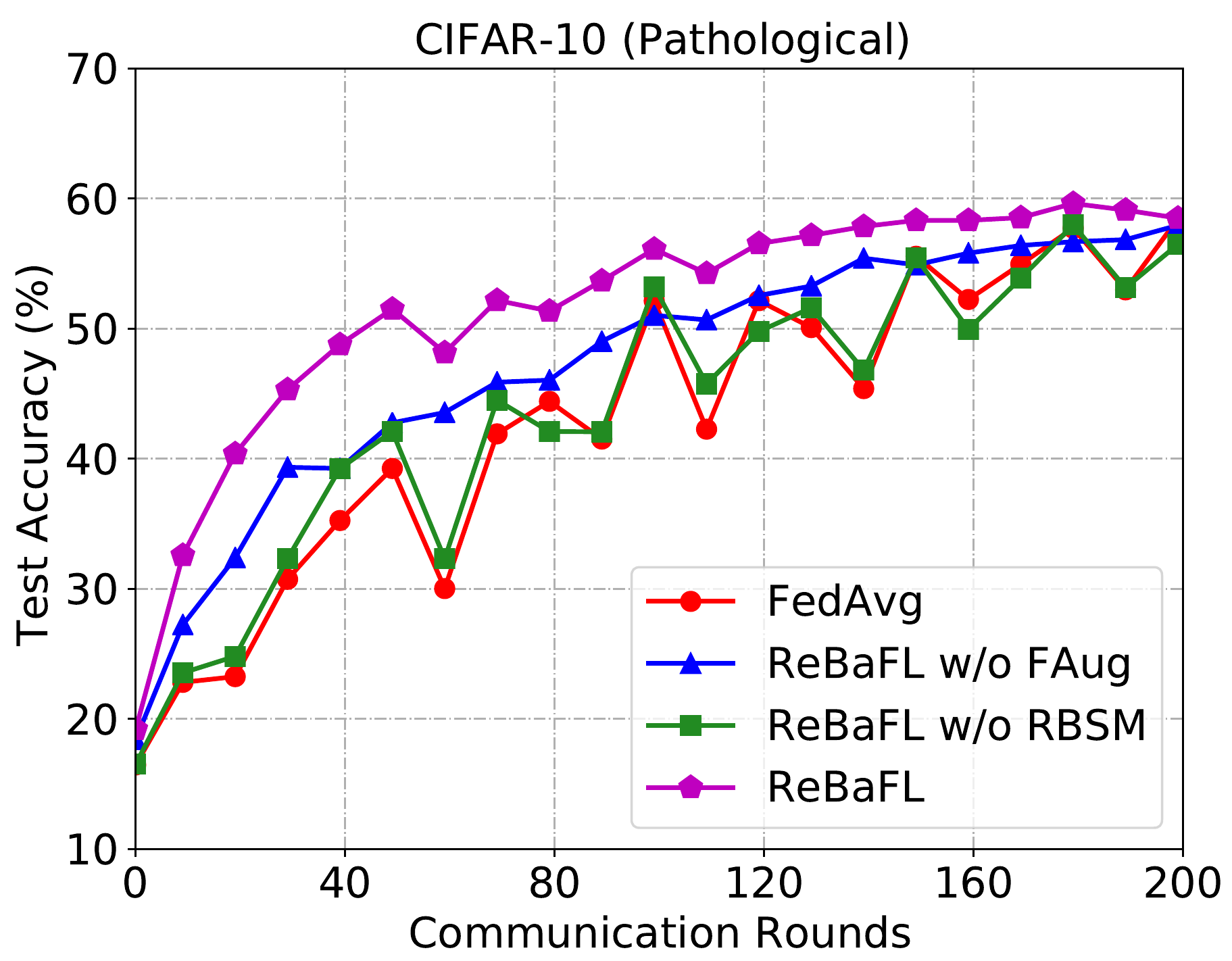}
	}
	%	\quad
	\subfigure[]{
		\includegraphics[width=0.46\columnwidth]{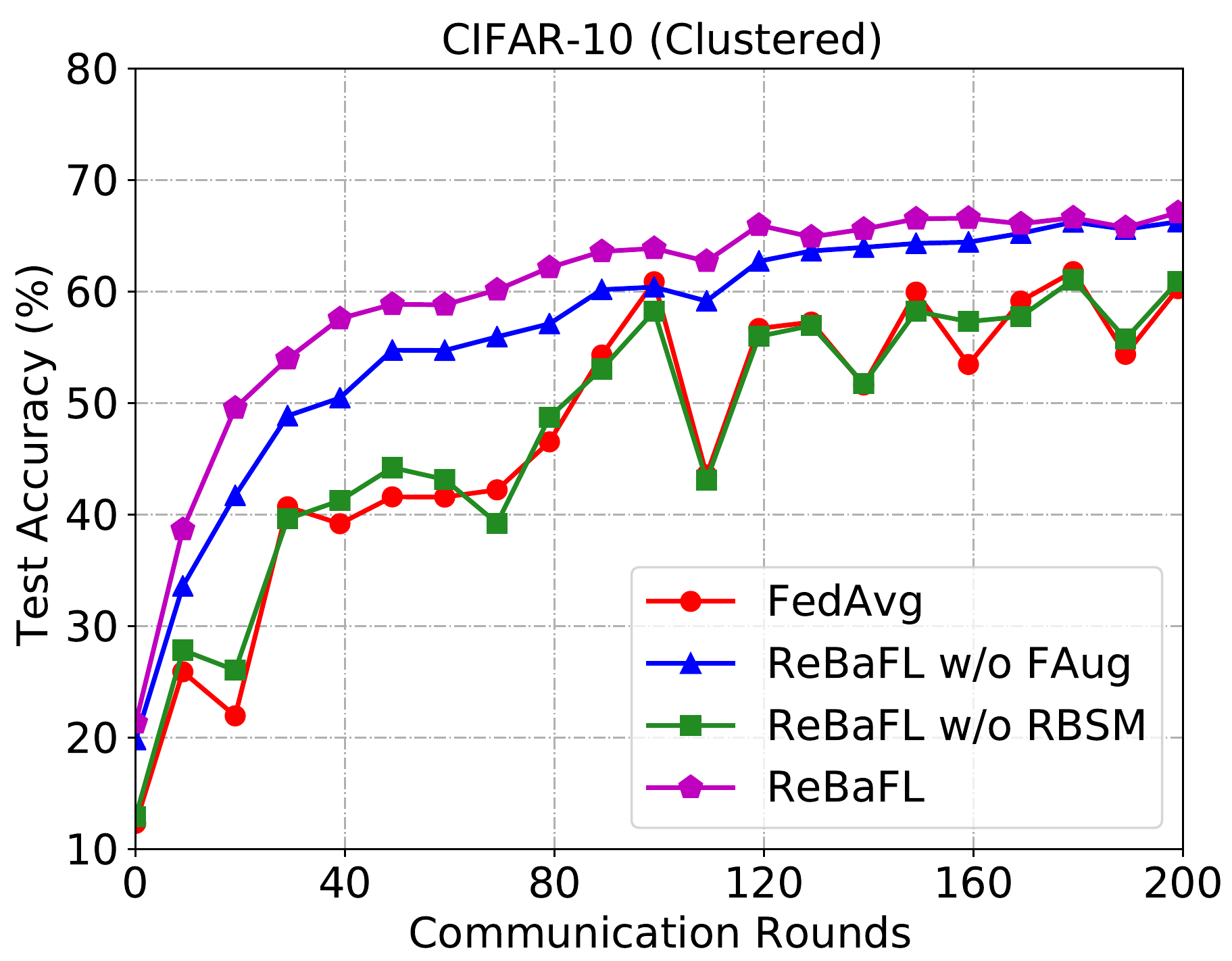}
	}
	%	\quad
	\caption{Learning curves for ablation studies. (a)(c) are in pathological setting on two datasets while (b)(d) are in clustered setting.}
	\label{fig:ablation}
		\vspace{-2ex}
\end{figure}

%\subsubsection{Efficacy of Classifier Training}

\subsubsection{Diversity of Local Models}
\begin{figure}[h]
	\centering
	\subfigure{
		\includegraphics[width=0.47\columnwidth]{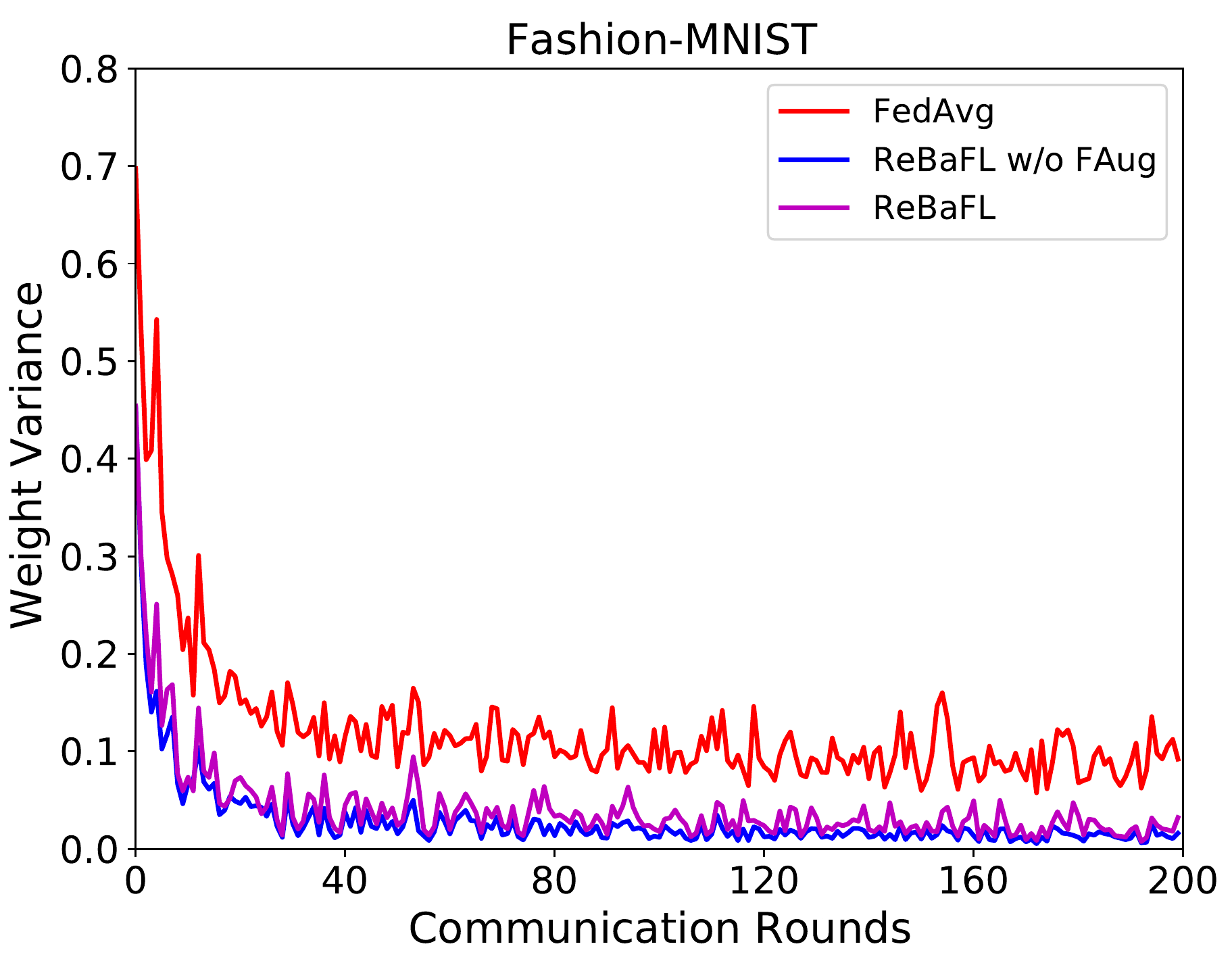}
	}
	%	\quad
	\hspace{-2ex}
	\subfigure{
		\includegraphics[width=0.47\columnwidth]{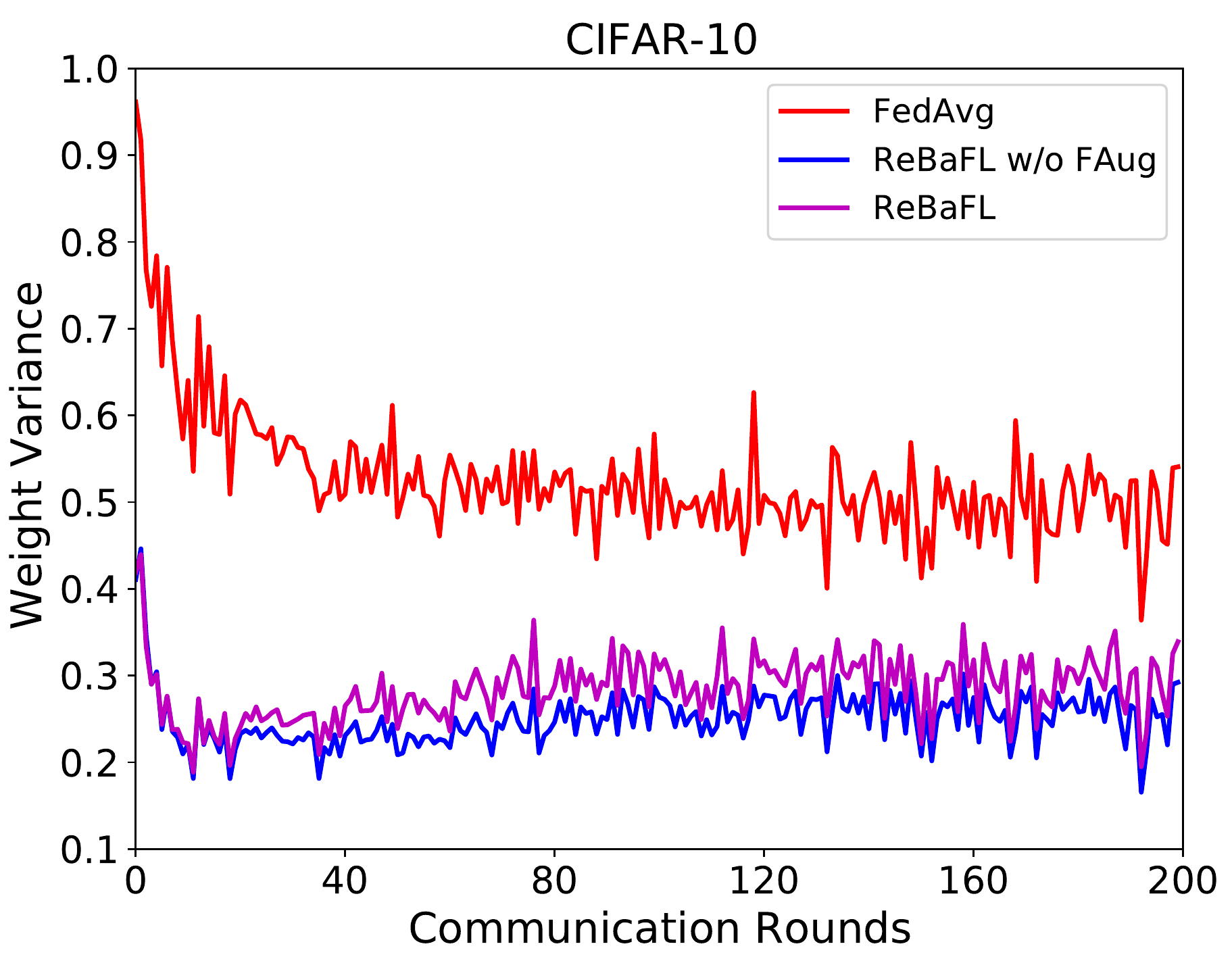}
	}
	%	\quad
	\caption{Local model parameter diversity over training rounds.}
	\label{fig:weight_var}
		\vspace{-2ex}
\end{figure}

We also record the model weight diversity among collected clients across the training process, where the weight diversity is defined as follows. 
\begin{equation}
\Lambda^{(t)} := \frac{1}{|S_t|} \sum_{i \in S_t} \|{\bm{w}}_{i}^{(t)} - \bar{\bm{w}}^{(t)}\|_{2}^{2},  
\end{equation}
where $\bar{\bm{w}}^{(t)}$ is the average of local models. The results in Fig.~\ref{fig:weight_var} indicate that our proposed balanced softmax can effectively reduce the weight diversity and thus facilitate the model aggregation, resulting in a better global model. Interestingly, we can find that when the feature augmentation is applied, the weight diversity will slightly increase as different clients may have various kinds of training samples and the calibration of local classifier also has some non-zero diversity across clients. However, the empirical results in the main evaluation clearly shows that the feature augmentation could benefit the federated learning by accelerating the model convergence and improving the final model accuracy. Possible explanation towards this phenomenon might be that slight noise in stochastic gradients can enhance the generalization of the trained model \cite{Smith20noise}.

%\subsubsection{Feature Embedding Visualization}

\subsection{Further Discussions}
As there are many variable factors in algorithm assessment, we present further analyses and discussions on some important aspects, including the choice of hyper-parameters, local training epochs, straggling behavior transition periods and system sizes.

\vspace{0.5ex}
\subsubsection{Impact of Hyper-parameters}
\begin{figure}[h]
	\centering
	\subfigure[]{
		\includegraphics[width=0.46\columnwidth]{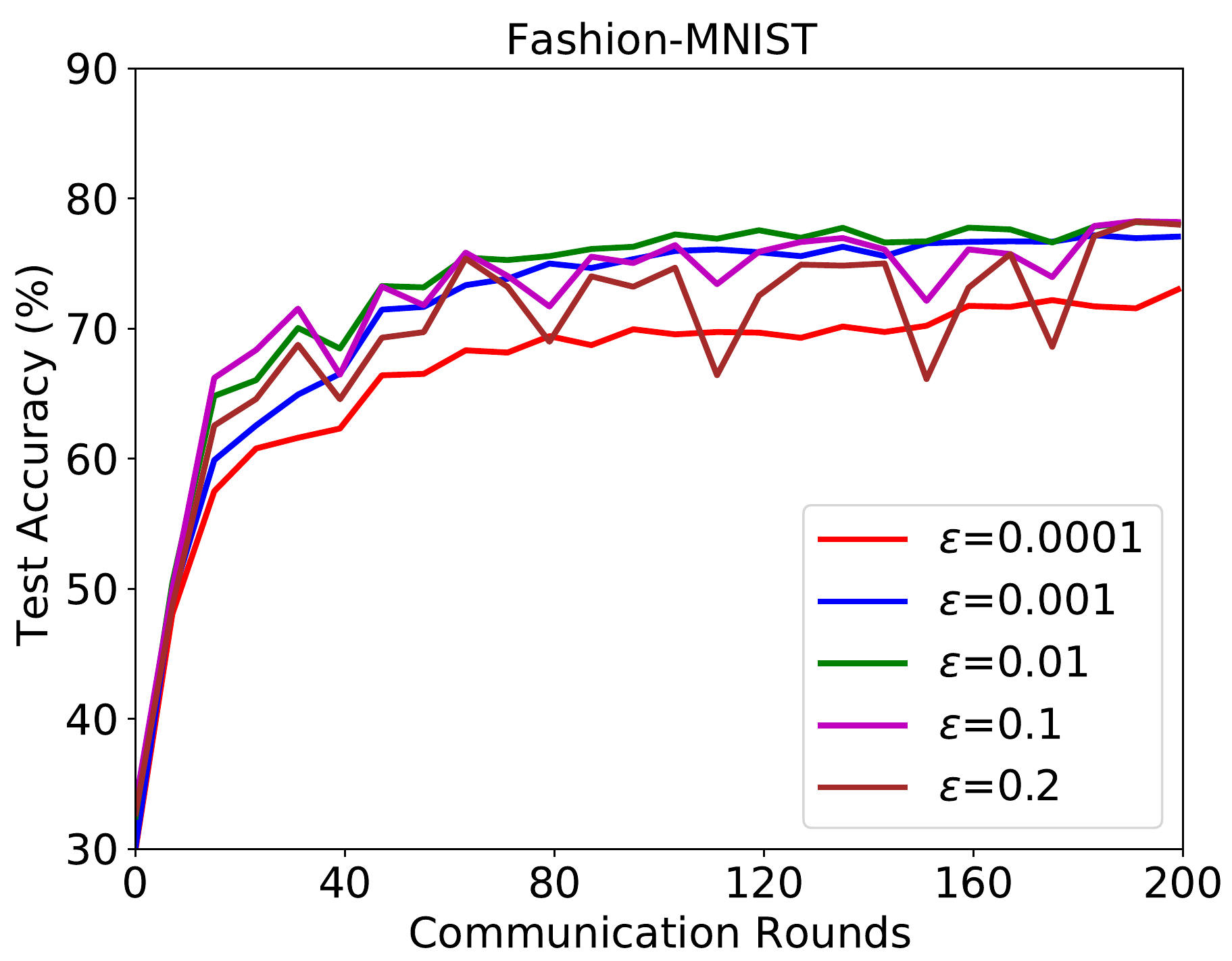}
	}
	%	\quad
	\subfigure[]{
		\includegraphics[width=0.46\columnwidth]{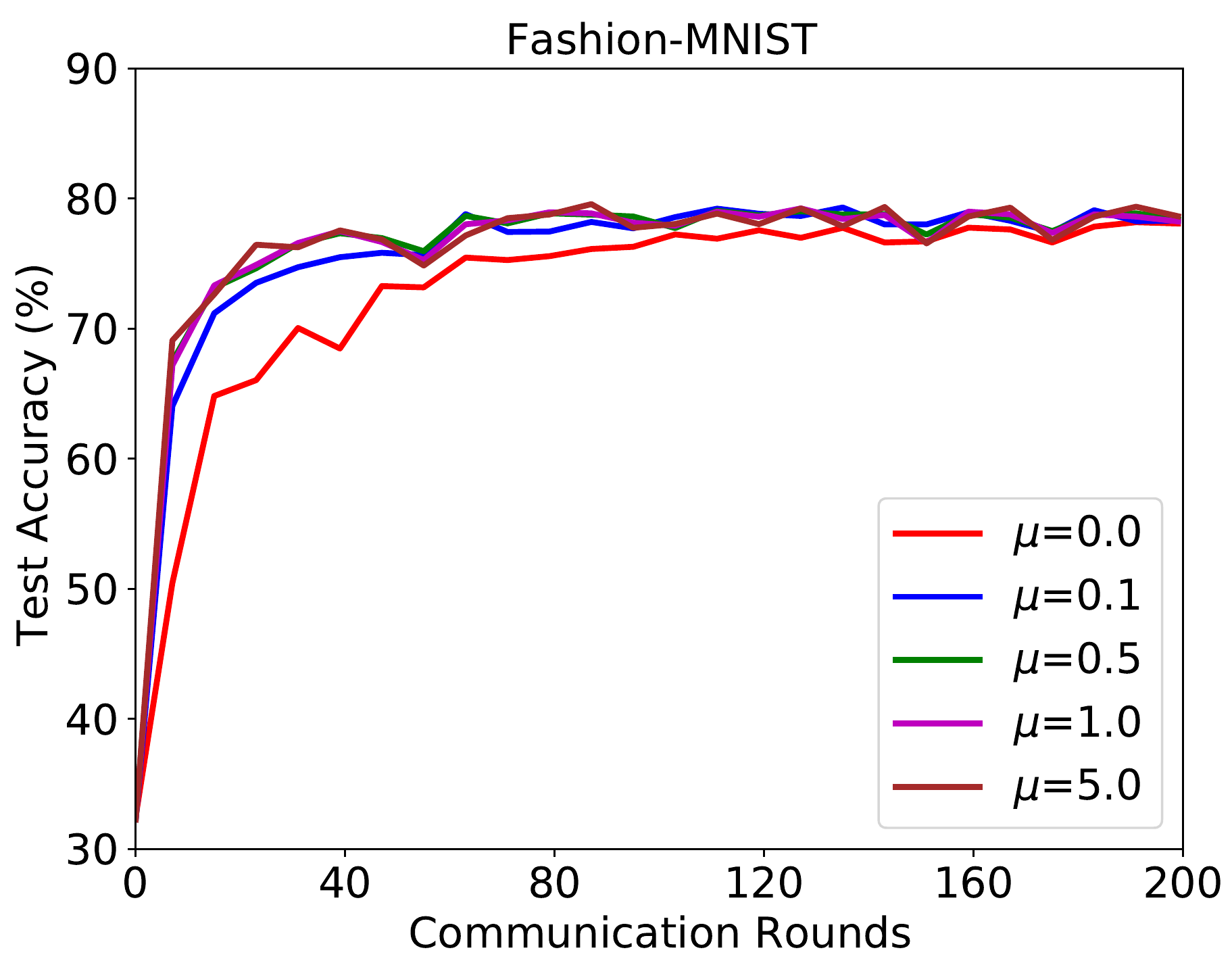}
	}
	%	\quad
	\subfigure[]{
		\includegraphics[width=0.46\columnwidth]{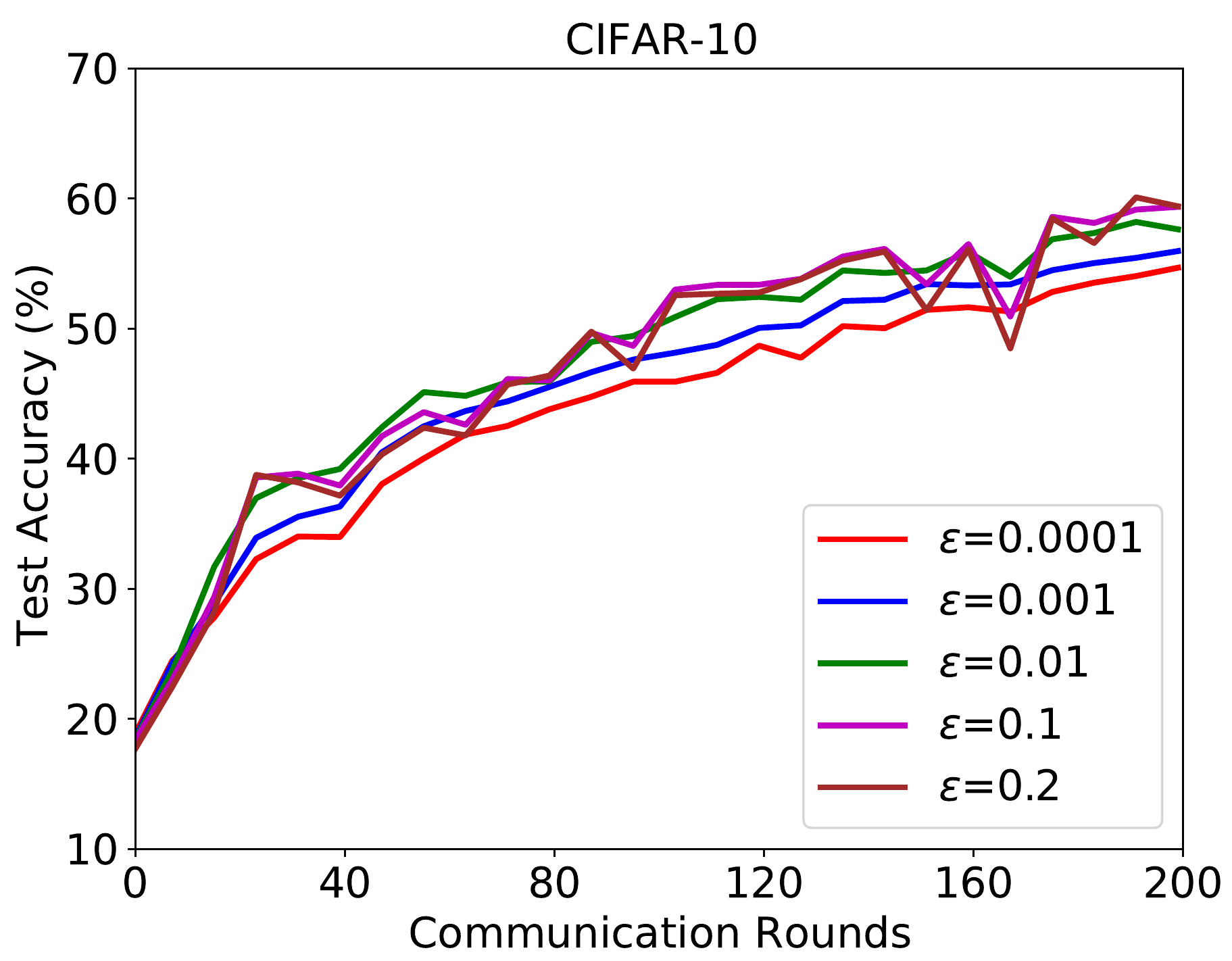}
	}
	%	\quad
	\subfigure[]{
		\includegraphics[width=0.46\columnwidth]{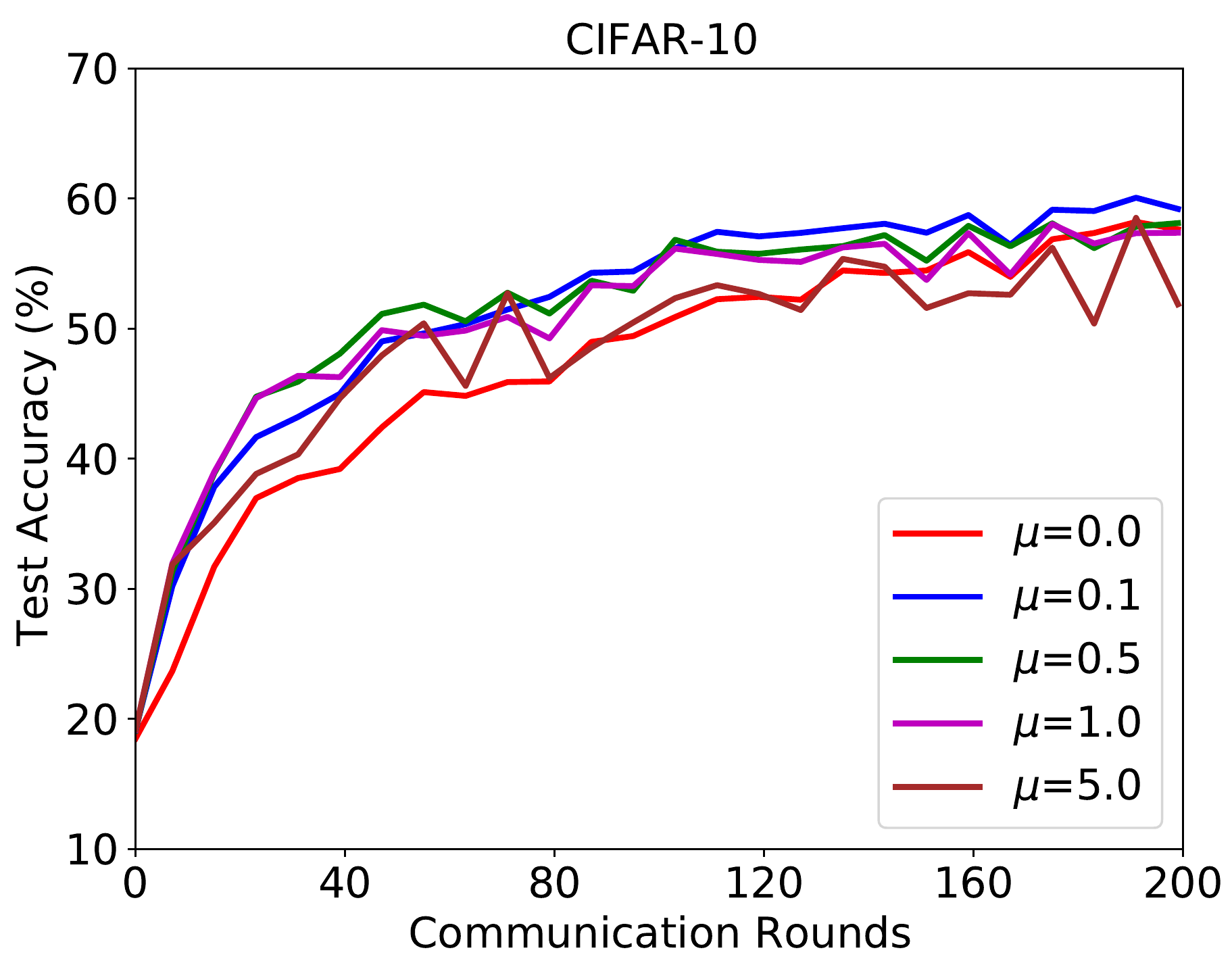}
	}
	%	\quad
	\caption{Performance comparison with varying hyper-parameters.}
	\label{fig:hypers}
	%	\vspace{-2ex}
\end{figure}

There are mainly two hyper-parameters ($\epsilon$ and $\mu$) in our framework, however, simultaneously tuning them is still time-consuming. Here we choose to incrementally add those two components by first investigating the $\epsilon$ to find a suitable value, then and tuning the $\mu$ with the selected $\epsilon^*$ fixed. Specifically, we vary the value of $\epsilon$ over $\{1e-4, 1e-3, {1e-2}, 1e-1, 2e-1\}$ with $\mu*=0$ and vary the value of $\mu$ over $\{0, 0.1, 0.5, 1.0, 5.0\}$ with $\epsilon^*=0.01$. The results are presented in Fig. \ref{fig:hypers}, from which we can find that a relative large $\epsilon$ will cause the training unstable but a too small value will reduce the model performance, indicating the importance of a suitable value of $\epsilon$. Similarly, a proper selection of $\mu$ is also needed. However, the best values of hyper-parameters may vary among specific learning tasks. As mentioned before, we do not aim at finding the best combination of hyper-parameters and the empirical studies demonstrate that relatively small values are preferred.

%\subsubsection{Impact of Server-Side Learning Rate}
%In the previous experiments, we set the server-side learning rate by $\eta_s=1$, however, as theorem shows that a properly larger  $\eta_s$ is beneficial for model convergence. to investigate the impact, we choose $\eta_s$ from $\{1.0, 1.2, 1.5, 1.8, 2.0, 3.0\}$ and compare BRFed with FedAvg, plotting the testing accuracy.

\subsubsection{Impact of Local Epochs}

\begin{figure}[t]
	\begin{center}
		\includegraphics[width=0.9\columnwidth,clip=true]{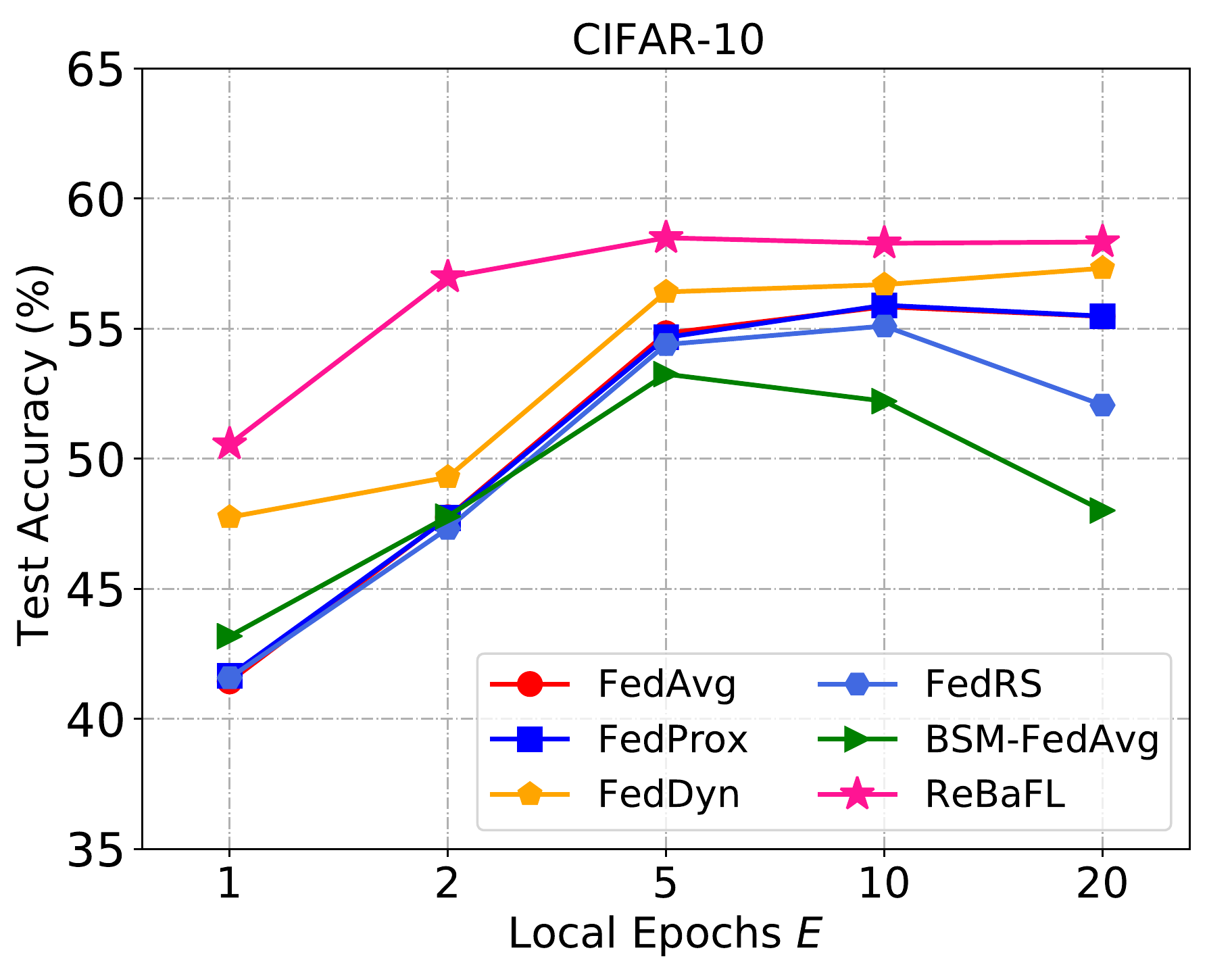}
	\end{center}
	\caption{Performance comparison on CIFAR-10 dataset with varying local epochs. The number of communication rounds is fixed.}
	\label{fig:local_epoch}
\end{figure}

In this part, we add more computation per client on each round by increasing the local epochs $E$. To reduce the communication overhead, the clients tend to have more local training steps for less frequent communication. To study the impact of local epochs, we conduct experiments on CIFAR-10 dataset with 20 clients by varying the epochs over $\{1,2,5,10,20\}$. As shown in Tab.~\ref{fig:local_epoch}, our method is superior to other baselines by a large margin. Given limited communication rounds, the model is not well trained for all methods when the number of local updates is too small. For a large value of $E$, while the performance gap of these baselines is relatively small, our approach still consistently outperforms other methods. The results also show that larger values of $E$ will only slightly hurt the model accuracy under our framework, while FedAvg results in more obvious accuracy degradation. It should also be pointed that the choice of  $E$ is associated to the trade-off among computation, communication and model performance.

\begin{figure}[t]
	\centering
	\subfigure{
		\includegraphics[width=0.47\columnwidth]{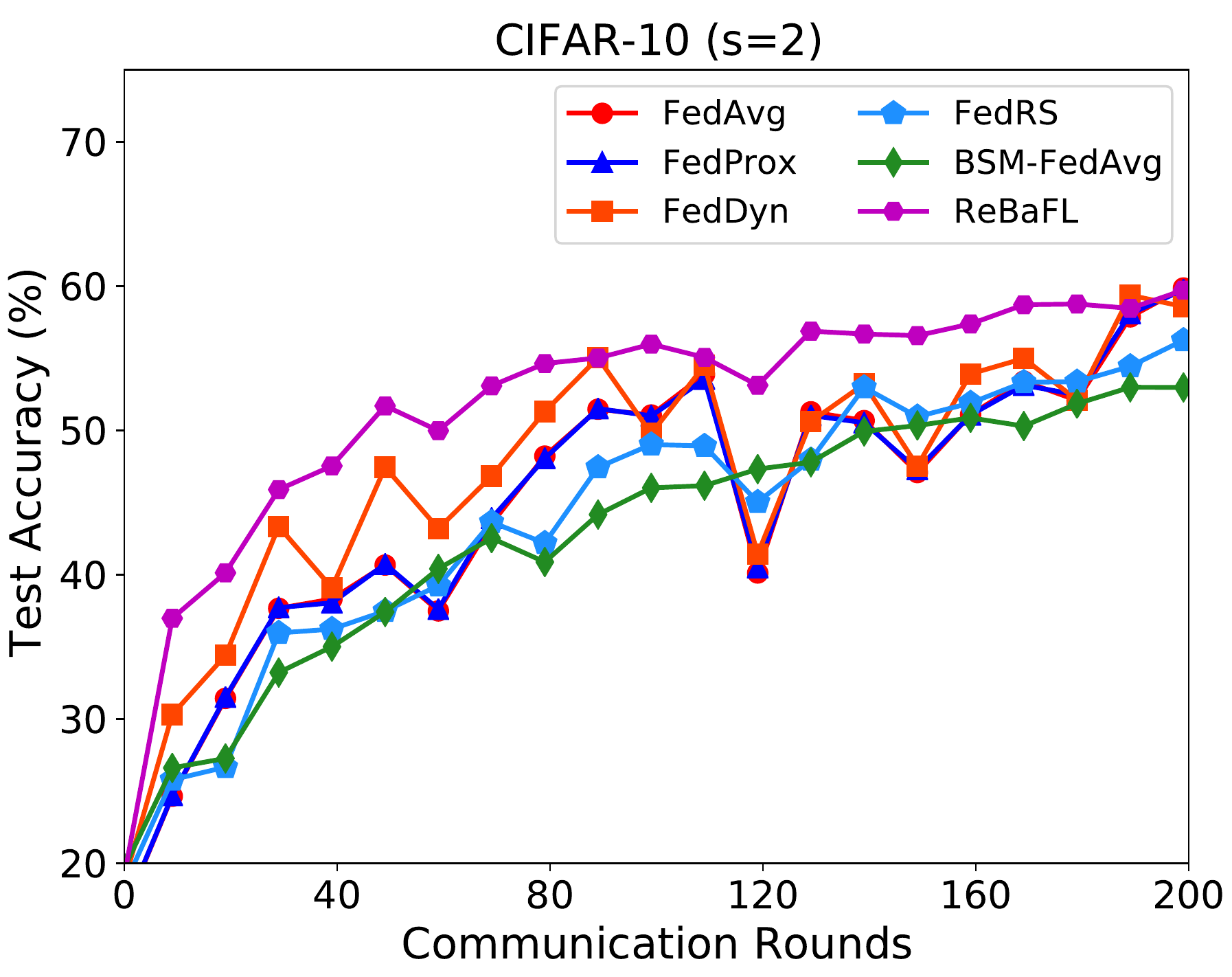}
	}
	%	\quad
	\hspace{-2ex}
	\subfigure{
		\includegraphics[width=0.47\columnwidth]{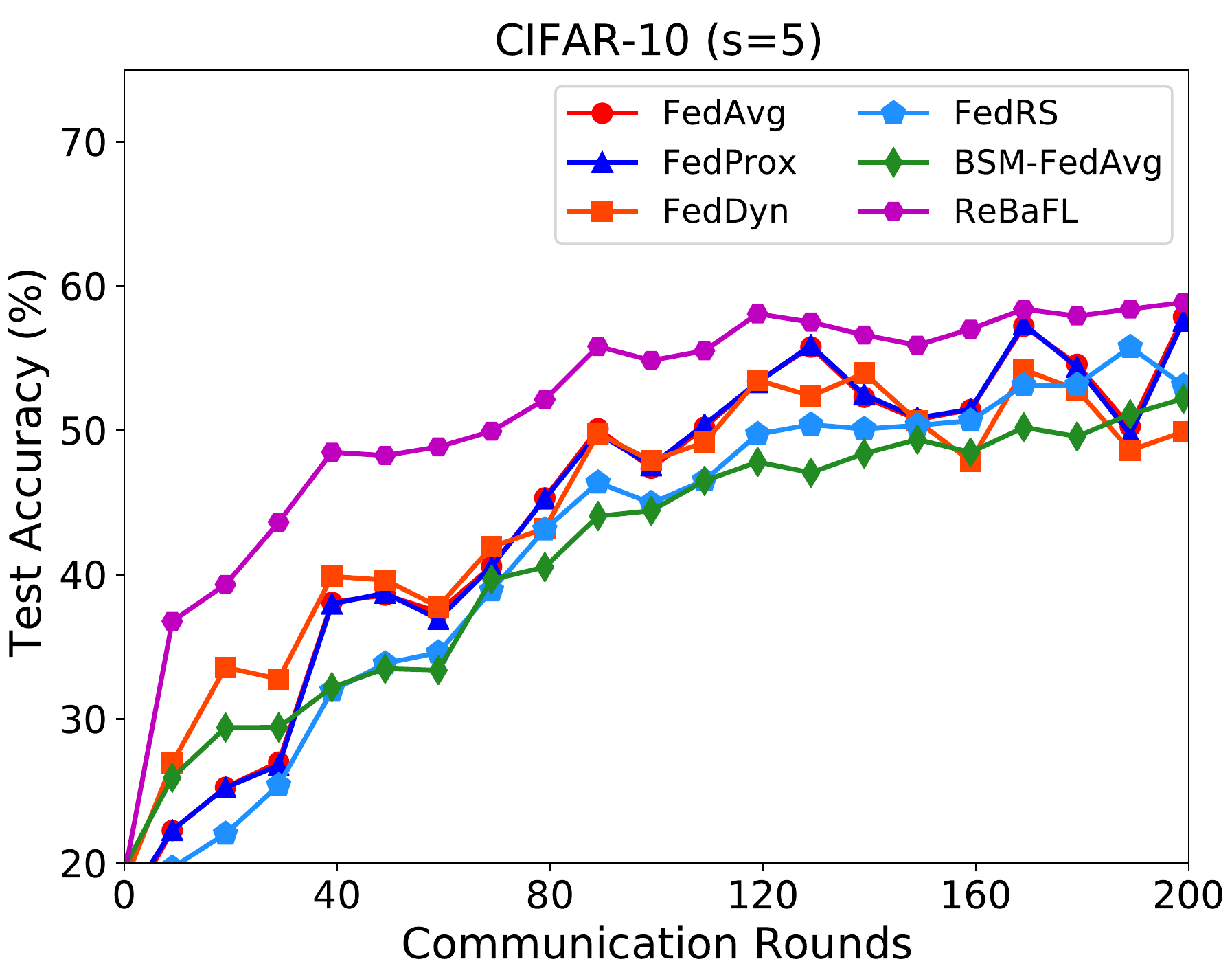}
	}
	%	\quad
%	\hspace{-2ex}
	\subfigure{
		\includegraphics[width=0.47\columnwidth]{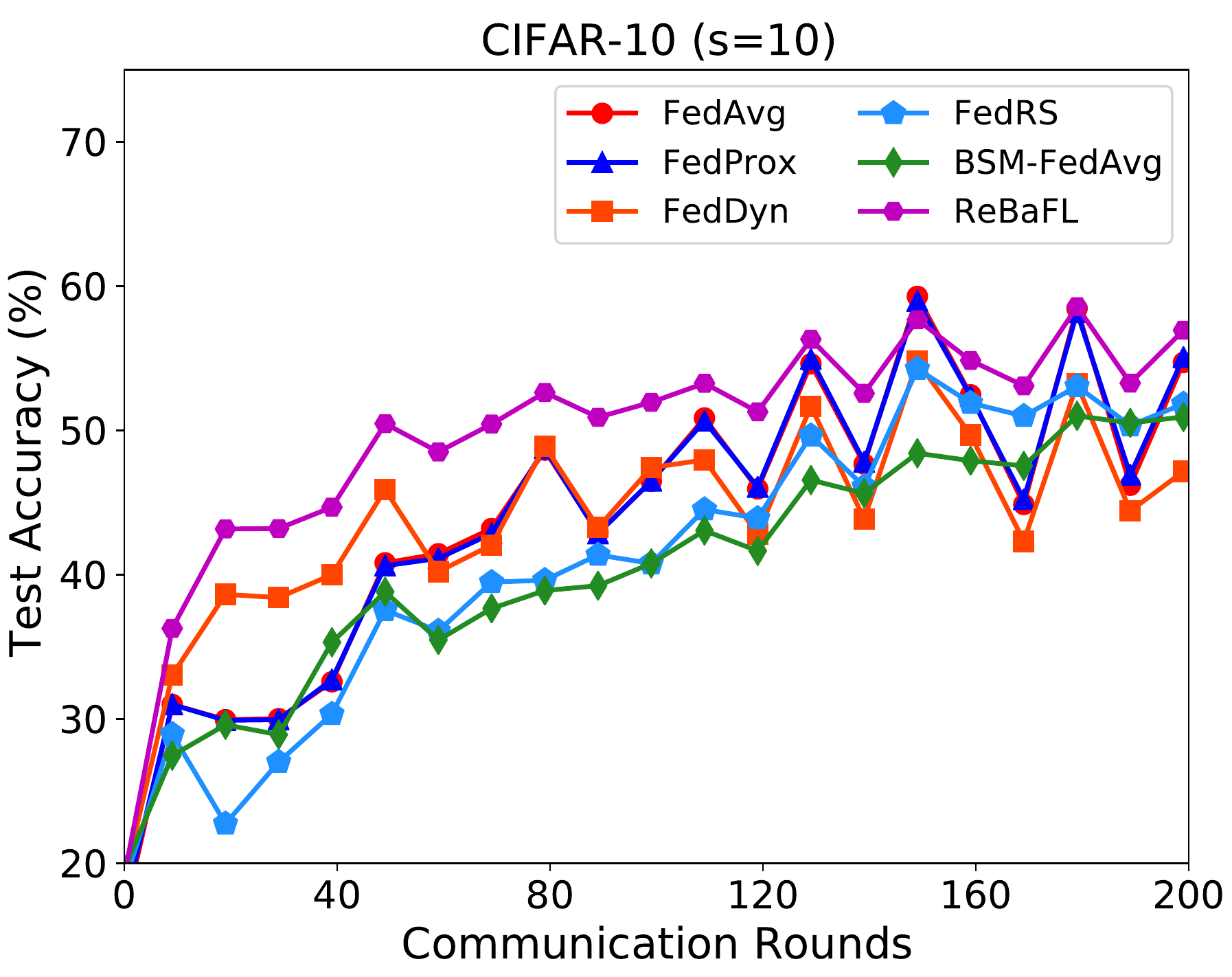}
	}
	%	\quad
	\hspace{-2ex}
	\subfigure{
		\includegraphics[width=0.47\columnwidth]{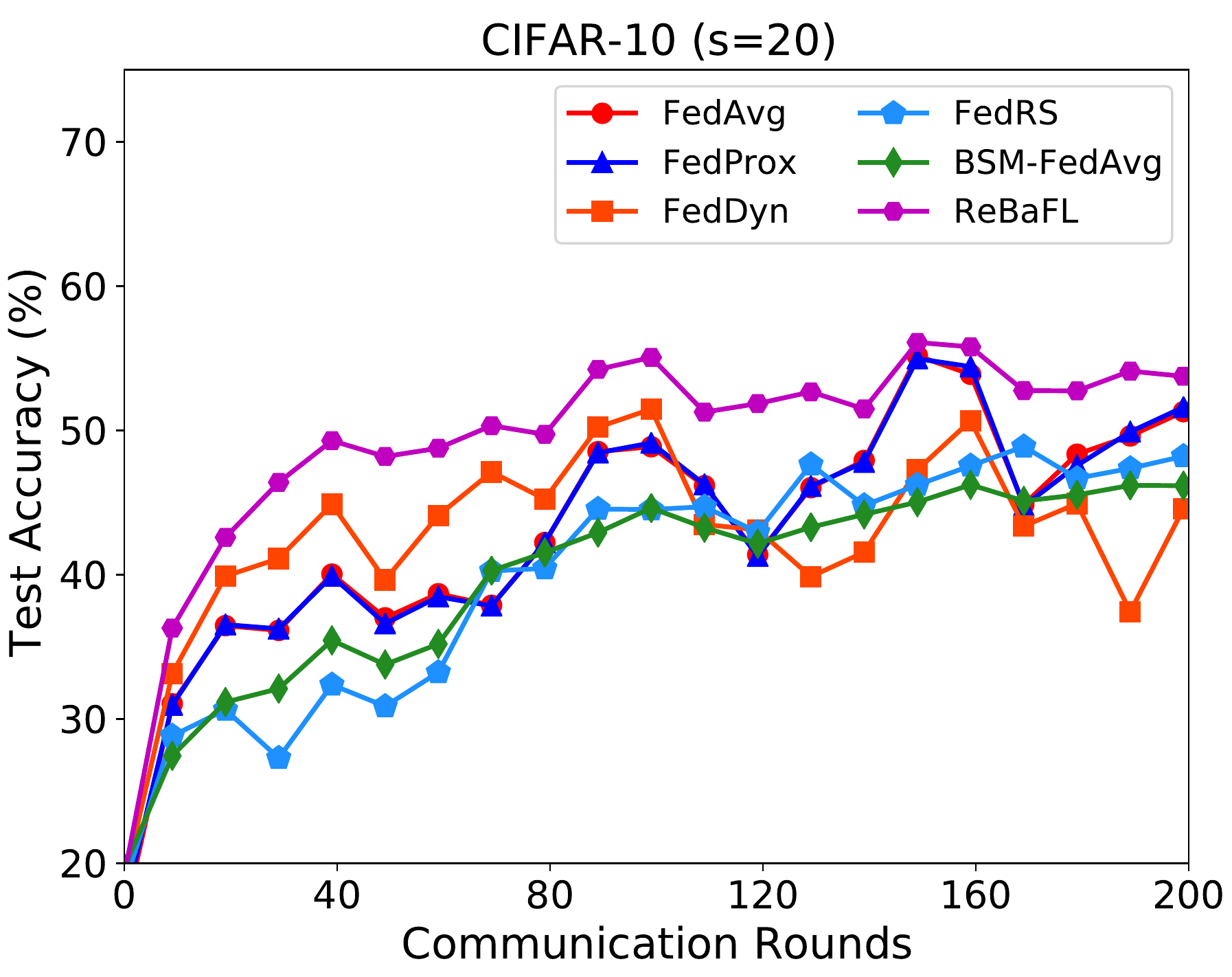}
	}
	%	\quad
	\caption{Performance comparison on CIFAR-10 dataset with varying straggling periods. The active status of each client is changed with probability $p$ at a interval of $s$ rounds.}
	\label{fig:straggling}
		\vspace{-2ex}
\end{figure}

\vspace{0.5ex}
\subsubsection{Impact of Straggling Behaviors}
In the above experiments, we always assume that the client dropout is occurred with a given probability and is independent across communication rounds, which however could only cover a limited scenario as a straggling or inactive client might persist for a long period so as for a active client. Therefore, we investigate the impact of persistent straggling behaviors by adjusting the active states with a varying period $s \in \{2,5,10,20\}$, where we conduct a random selection of dropout clients at intervals of every $s$ rounds. From the curves in Fig.~\ref{fig:straggling}, it can be found that the straggling period has a non-negligible impact on the federated learning as in each period some classes may be missing among the active clients and the learned global model is biased such that the test accuracy might decrease in a extent. Nevertheless, our method still exhibits the best performance across different straggling behaviors, demonstrating its robustness and stability against the client dropout in cross-device FL scenarios.

\vspace{0.5ex}
\subsubsection{System Scalability} In realistic IoT systems, the number of clients could be large, therefore it is important to evaluate the adaptability and scalability of FL algorithms. To this end, we launch three systems with $m$=100, 200 and 500 clients, respectively. We evenly partition the whole CIFAR-10 dataset into  $m$ clients and ensure that each client only contains samples from 2 classes. For simplicity, we select 10\% clients in each communication round at random to simulate the independent stragglers. The results presented in Table~\ref{tab:scalability} demonstrate that our proposed ReBaFL still outperforms other baselines under different system sizes. It can also be observed that the performance of all methods decreases with $m$ increasing. We conjecture that the reason is that more clients will make the local models more diverse and hurt the global aggregation. However, our method can still achieve 46.67\% accuracy when there are 500 clients.
%\vspace{-1ex}

\begin{table}[h]
	\small
	\caption{The test accuracy (\%) with varying number of clients ($m$) on CIFAR-$10$. Client sampling rate is 0.1 for all settings.}
	\centering
	
	\begin{tabular}{p{2.5cm} ccc}
		\toprule
		\multirow{2}{*}{\textbf{Method}}& \multicolumn{3}{c}{\textbf{System Size}}\\
		\cmidrule(lr){2-4}
		& {$m=100$} & {$m=200$} & {$m=500$} \\
		\midrule
		FedAvg
		& 55.78 & 47.53 & 33.78
		\\
		FedProx
		& 56.81 & 47.49 & 33.81
		\\
		Scaffold
		& 63.36 & 47.66 & 16.12
		\\
		FedDyn
		& 63.35 & 53.58 & 37.74
		\\
		MOON
		& 50.53 & 42.47 & 26.89
		\\
		FedRS
		& 60.85 & 51.86 & 42.12
		\\
		FedBABU
		& 52.65 & 43.62 & 33.41
		\\
		\midrule
		BSM-FedAvg
		& 60.32 & 51.75 & 43.36
		\\
		\textbf{ReBaFL} (ours)
		& \textbf{63.89} & \textbf{57.23} & \textbf{46.67}
		\\
		\bottomrule
	\end{tabular}
	\label{tab:scalability}
\end{table}

\begin{figure*}[h]
	\begin{center}
		\includegraphics[width=1.96\columnwidth,clip=true]{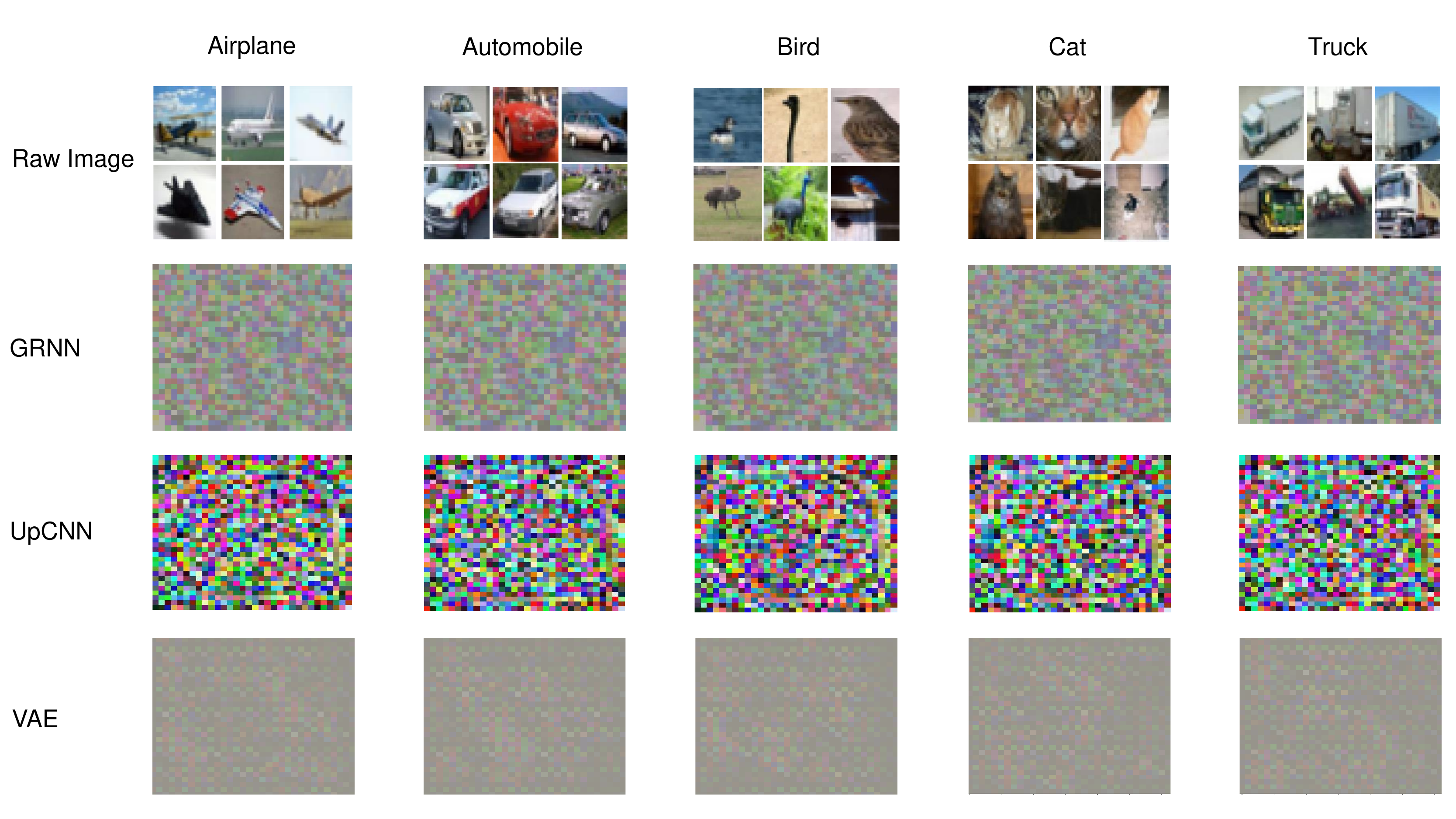}
	\end{center}
	\vspace{-1ex}
	\caption{Reconstruction results of invert attacks on the feature prototypes. No semantically meaningful information can be extracted by the attacks.}
	\label{fig:invert}
\end{figure*}

\vspace{-2ex}

\subsection{Testing Data Leakage from Prototypes}
Finally, we study the potential privacy leakage risk during the training of ReBaFL. Since our training process is involved with the sharing of feature prototypes for each class, we consider the honest-but-curious setting where both server and clients follow the system protocol exactly as specified, but may try to infer other parties’ private information based on the received feature prototypes. We simulate the scenarios with possible information leakage risks by presenting several data leakage attacks on CIFAR-10 dataset, including up-convolutional neural network (UpCNN) \cite{DosovitskiyB16} and variational autoencoder (VAE) \cite{Kingma14VAE} that try to recover the input images from the feature representations. And we also adopt the approach from the state-of-the-art gradient attack generative regression neural network (GRNN) \cite{Ren22GRNN} to inverse the feature prototypes. From the reconstructed results in Fig.~\ref{fig:invert}, we can conclude that sharing feature prototypes will neither leak considerable information about the raw data, nor will they reveal any relationship between different image categories. {The reasons why feature prototypes are difficult to invert may be in three-fold. First, the private information contained by the low-dimensional feature representation is much less than the high-dimensional gradient, thus even the successful gradient attack might fail to infer raw data from feature prototypes. Second, the feature prototype is an average of per-image feature vectors and the characteristics of individual images could be concealed. Third, the numerous non-linear transformations in the hidden layers of complex neural networks make it hard to invert images by only aligning the feature representations.} This is also consistent with the phenomenon that reconstruction from the shallow representations is easier than from the representations generated by the deep fully-connected layers \cite{DosovitskiyB16}.

\vspace{-1ex}

\section{Conclusion}\label{secConclusion}

In this paper, we introduce the relaxed balanced-softmax and the cross-class transfer-based feature augmentation for stabilizing and enhancing the federated learning with non-IID data and client dropout. We provide extensive evaluations to demonstrate the non-trivial improvement achieved by of our framework. We also evaluate image invert attacks on the feature prototypes to verify that the sharing of prototypes will not induce high privacy leakage risks. Promising future directions include extending our framework to the asynchronous/decentralized systems and incorporating more advanced global aggregation strategies to integrate heterogeneous local models.

\vspace{-1ex}

%
%\section*{Acknowledgment}\label{secAcknowledgment}
%The research of Shao-Lun Huang is supported in part by the Shenzhen Science and Technology Program under Grant KQTD20170810150821146, National Key R\&D Program of China under Grant 2021YFA0715202.

%\newpage
\bibliographystyle{IEEEtran}
\bibliography{mybib}
%\clearpage
%\input{10.appendix}
%\newpage

\end {document}